\documentclass[conference,compsoc]{IEEEtran}

\ifCLASSOPTIONcompsoc
  \usepackage[nocompress]{cite}
\else
  \usepackage{cite}
\fi

\usepackage[pdftex]{graphicx}
\graphicspath{{figures/}}
\ifCLASSOPTIONcompsoc
  \usepackage[caption=false,font=footnotesize,labelfont=sf,textfont=sf]{subfig}
\else
  \usepackage[caption=false,font=footnotesize]{subfig}
\fi
\usepackage{dblfloatfix}

\usepackage{amssymb}

\usepackage{xurl}
\usepackage[most]{tcolorbox}
\tcbset{
  hylist/.style={
    enhanced,
    colback=gray!5,
    colframe=black,
    boxrule=0.5pt,
    arc=2pt,
    left=6pt,
    right=6pt,
    top=4pt,
    bottom=4pt,
    boxsep=2pt,
  }
}

\usepackage[hidelinks]{hyperref}

\usepackage{multirow}
\usepackage{booktabs}
\usepackage{caption}
\usepackage{makecell}
\usepackage{tabularx}
\usepackage{array}
\newcolumntype{M}[1]{>{\centering\arraybackslash}m{#1}}
\newcolumntype{P}[1]{>{\centering\arraybackslash}p{#1}}
\newcolumntype{Y}{>{\centering\arraybackslash}X}

\usepackage{float}
\usepackage[available,functional,reproduced]{ieeebadges}

\begin{document}

\title{From GAN-Based Models to GPT-5: A Taxonomy-Guided Study of Human Performance in Detecting Deepfake Face Images}

\author{\IEEEauthorblockN{Zhiqing Guo\IEEEauthorrefmark{1}\IEEEauthorrefmark{6}, Lixin Jia\IEEEauthorrefmark{1}\IEEEauthorrefmark{6}, Luning Sun\IEEEauthorrefmark{2}\IEEEauthorrefmark{6}, Xiaowei Chen\IEEEauthorrefmark{3}, Guofeng Yu\IEEEauthorrefmark{1}\protect\\
Liejun Wang\IEEEauthorrefmark{1}\IEEEauthorrefmark{7}, Gaobo Yang\IEEEauthorrefmark{4}, David Stillwell\IEEEauthorrefmark{2}, Thorsten Holz\IEEEauthorrefmark{3} and Shujun Li\IEEEauthorrefmark{5}\IEEEauthorrefmark{7}}
\IEEEauthorblockA{\IEEEauthorrefmark{1}Xinjiang University, China
\quad \IEEEauthorrefmark{2}University of Cambridge, UK}
\IEEEauthorblockA{\IEEEauthorrefmark{3}Max Planck Institute for Security and Privacy (MPI-SP), Germany
\quad \IEEEauthorrefmark{4}Hunan University, China}
\IEEEauthorblockA{\IEEEauthorrefmark{5}Institute of Cyber Security for Society (iCSS), University of Kent, London and South East University Group, UK}
\IEEEauthorblockA{\IEEEauthorrefmark{6}Equivalent contribution
\IEEEauthorrefmark{7}Corresponding co-authors: wljxju@xju.edu.cn (Liejun Wang), S.J.Li@kent.ac.uk (Shujun Li)}}

\maketitle

\begin{abstract}
Generative models for deepfakes have rapidly evolved from GAN-based architectures to diffusion models and multimodal large language models (MLLMs). However, empirical research on human performance in detecting deepfake face images has not kept pace. It remains unclear if (and how) detection performance varies across generative models and whether model-specific training improves detection more effectively than general-purpose training. In this paper, we propose the \emph{Deepfake Face Artifact Taxonomy} (DFAT), a cross-model taxonomy of human-recognizable visual artifacts in deepfake face images, based on which we build both general-purpose and model-specific training materials. Through an online experiment ($n = 524$), we examined human performance in detecting deepfake face images and evaluated training efficacy across five representative generative models: three GAN-based, one diffusion-based, and one MLLM (GPT-5). Our results show that participants’ detection accuracy was strongly model-dependent and that images generated by more recent and advanced models were not necessarily harder to detect. After training, participants achieved significant performance gains across most conditions, and their improvement varied by model. Notably, model-specific training generally yielded greater numerical gains than general-purpose training. In a five-month follow-up test ($n = 342$), we observed no significant decline in detection accuracy and no significant increase in perceived stress. Overall, our findings provide actionable insights for designing effective, sustainable, and targeted training programs to counter the growing threat posed by deepfakes.
\end{abstract}

\section{Introduction}

Deepfake face images are frequently exploited for malicious purposes, including disinformation campaigns~\cite{Vaccari-Chadwick:DeepfakeDisinfo:SMS2020}, non-consensual pornography~\cite{Furizal:DeepfakePorn:SSHC2025, Li:AIGNCII:CHI2026}, online fraud~\cite{Lin:DeepfakeFraud:FiL2025}, and cyberbullying~\cite{Alexander:DeepfakeCyberbullying:CH2025}. The recent development of generative AI techniques, particularly multimodal large language models (MLLMs)~\cite{Caffagni:MLLMsurvey2024, Yin:MLLMsurvey2024} and large vision models (LVMs)~\cite{Zhang:LVMsurvey2024}, has significantly escalated such threats. These modern deepfake generation models have lowered the technical barriers substantially, allowing adversaries to generate highly realistic deepfakes from textual and multimodal prompts, without the need for extensive expertise or computational resources.

Unfortunately, existing human-focused studies have not kept pace with the rapid development of modern deepfake generators. Most prior work on deepfake face images relies on relatively old GAN-based architectures, particularly StyleGAN2~\cite{Karras-StyleGAN2-CVPR2020}. Only one recent study~\cite{Lu:seeing2023} examined diffusion-based image generators (e.g., Midjourney V5~\cite{Midjourney}) without considering any MLLMs. Cooke et al.~\cite{Cooke:Asgood2025} included both GAN-based and diffusion models, but failed to report model-specific detection performance. Without empirical evidence, it remains unclear whether and how human performance in detecting deepfake face images varies across different generative models.

Current approaches to mitigating deepfakes focus predominantly on automated detection~\cite{Mirsky-Lee:DeepfakeSurvey2021, Dixit:DeepfakeAudioSurvey2023, Pu:DeepfakeTextDetection:SP2023, Uchendu:DeepfakeTextsTutorial:NAACL2024}. However, detector performance often degrades under distribution shifts to unseen generators and adversarial manipulation~\cite{Li:DeepfakeSpeechSurvey2025, Wang:DeepfakeDetectionSurvey2025}. These tools may also be unavailable in real-time or end-to-end encrypted settings. Even when they are available, users still need to determine whether they should trust these outputs and how to act on them. As a result, users are left to assess image authenticity on their own, making user-facing support for deepfake discernment an emerging topic in human-centered security~\cite{Geissler:DeepfakeLiteracy:CHI2026}.

Despite this need, human performance in detecting deepfakes remains comparatively understudied. A 2024 meta-analysis~\cite{Alexander-human-CHBR2024} identified 56 studies spanning multiple modalities (audio, image, text, and video). Constraining the scope to human-subject studies of deepfake face images, we identified only nine studies that reported quantitative performance for photorealistic facial deepfakes. Overall, these studies suggest that untrained users are poorly equipped to identify high-quality deepfakes, often performing at a near chance level~\cite{Frank-DeepfakeHumanDetection-SP2024, Shen:study2021}. Although some studies reveal that feedback-based training and tutorials can improve human performance~\cite{Boyd:value2023, Hulzebosch:DetectingCNN-generated2020, Diel:Inabilitytodetect2024, Nightingale-Farid:PNAS2022}, their effectiveness varies considerably across study designs, with several studies reporting largely negative results~\cite{Bray-testinghuman-JC2023}.

To address these gaps, we evaluate human performance in detecting deepfake face images generated by five representative models that span the current face-synthesis trajectory: StyleGAN2, StyleGAN3 and GRAM-HD as GAN-based face generators, SD3.5 as an open-weight diffusion model, and GPT-5 as a widely accessible MLLM-based generator. To improve training effectiveness, we construct programs that are grounded in a systematic synthesis of human-recognizable visual artifacts across these generative models. In this study, we aim to address the following research questions (RQs):\\
\indent \textbf{RQ1:} What human-recognizable visual artifacts characterize deepfake face images generated by different generative models?\\
\indent \textbf{RQ2:} Does human performance in detecting deepfake face images vary across different generative models, and if so, how?\\
\indent \textbf{RQ3:} How effective are taxonomy-guided training interventions in improving human performance in detecting deepfake face images, and is model-specific training more effective than general-purpose training?\\
\indent \textbf{RQ4:} Is the performance improvement derived from the taxonomy-guided training interventions sustainable over a relatively long period of time?

This study focuses on photorealistic human face images, one of the most prevalent and socially consequential forms of deepfakes. We begin by examining real and deepfake face images generated by different models. Building on this systematic analysis, we propose the \textit{Deepfake Face Artifact Taxonomy} (DFAT), which offers a structured and extensible framework for characterizing model-specific visual artifacts that can aid in deepfake detection. Guided by this taxonomy, we design two types of deepfake detection training interventions: general-purpose and model-specific. We then evaluate and compare their effectiveness through an online experiment comprising a main study ($n = 524$) and a follow-up study five months later ($n = 342$) assessing retention effects. This study makes the following key \textit{contributions}:

\begingroup
\setlength{\IEEEilabelindent}{0pt}
\begin{itemize}
\item We propose DFAT, an extensible taxonomy of human-recognizable artifacts in deepfake face images generated by diverse models. DFAT supports systematic comparison of artifact profiles across deepfake generation models and provides a structured basis for developing training interventions.

\item We present a cross-model baseline for human performance in deepfake detection and show that deepfakes generated by more recent and advanced models are not necessarily harder for humans to detect. This finding calls for a shift of future studies from focusing on more recent and advanced models to model-specific human-recognizable artifacts.

\item We develop DFAT-guided training interventions and demonstrate their effectiveness in improving deepfake detection accuracy, with model-specific training outperforming general-purpose training. These results offer practical guidance for designing future educational interventions for deepfake detection.

\item We provide longitudinal evidence on the retention of humans' deepfake detection performance through a follow-up test five months after the main experiment. The results reveal persistent improvement in detection accuracy and user confidence without increasing perceived stress, highlighting the sustainability of DFAT-guided training interventions.
\end{itemize}
\endgroup

\section{Related Work}

\begin{table*}[!htb]
\centering
\small
\caption{Description of the five generative models used in our study.}
\label{table:deepfake_models}
\begin{tabularx}{\linewidth}{P{11.3em} P{9.2em} c X}
\toprule
\textbf{Model} & \textbf{Venue/Vendor (Year)} & \textbf{Output Size(s)} & \textbf{Reason(s) of Selection}\\
\midrule
StyleGAN2~\cite{Karras-StyleGAN2-CVPR2020} & CVPR (2020) & $1024 \times 1024$ & The most widely studied high-fidelity GAN-based model\\
\midrule
StyleGAN3~\cite{Karras-StyleGAN3-NeurIPS2021} & NeurIPS (2021) & $1024 \times 1024$ & The most recent variant of the StyleGAN family with improved alias-free generation\\
\midrule
GRAM-HD~\cite{Xiang-GRAM-HD-ICCV2023} & ICCV (2023) & $1024 \times 1024$ & A GAN-based model more recent than StyleGAN3\\
\midrule
Stable Diffusion 3.5 Large (SD3.5 for short)~\cite{SD3.5-news, SD3.5-large} & Stability AI (2024) & Multiple sizes & The latest version of the Stable Diffusion open-weight text-to-image models supporting prompt-based generation\\
\midrule
GPT image generation (GPT-5 interface)~\cite{GPT-5} & OpenAI (2025) & Multiple sizes & One of the best-performing and widely used MLLMs when we started our work in mid 2025\\
\bottomrule
\end{tabularx}
\end{table*}

\subsection{Deepfake Generation Models}

Deepfake image generation now covers faces, objects, and scenes across several model families. GANs established high-fidelity image synthesis, with fully generative face synthesis becoming one of their most mature applications~\cite{Kammoun:GANsurvey2022, karras-PGGAN-2018}. Diffusion models and latent diffusion models made text-guided image generation easier to use widely~\cite{ROmbach-LDM-CVPR2022, Dustin:SDXL2024}. Recent systems, including MLLM-based interfaces and commercial image generators~\cite{Midjourney, Ramesh:DALL-E2022}, further allow users to generate realistic images using natural language inputs.

Human faces are a demanding target for image generation because they contain complex geometry, fine-grained textures, and rich expression variations. Face images also play a central role in social media, identity presentation, and online security, making realistic face synthesis one of the most visible forms of deepfake images. Deepfake face generation includes face manipulation, such as face swapping and editing~\cite{Chen:SimSwap2020, Choi:stargan2018}, as well as fully generative face synthesis. For fully generative face synthesis, the StyleGAN series remains a standard benchmark~\cite{Karras-FFHQ-CVPR2019, Karras-StyleGAN2-CVPR2020, Karras-StyleGAN3-NeurIPS2021}, and recent GAN-based approaches further improve visual quality and realism through new architectures and training strategies~\cite{Chai-Anyres-GAN-ECCV2022, Chan-3DGAN-CVPR2022, Xiang-GRAM-HD-ICCV2023}.

To operationalize this landscape for our empirical analysis, we select five representative deepfake models that cover the face-synthesis design space. We include StyleGAN2~\cite{Karras-StyleGAN2-CVPR2020} and StyleGAN3~\cite{Karras-StyleGAN3-NeurIPS2021} as widely used GAN-based face synthesis baselines, and GRAM-HD~\cite{Xiang-GRAM-HD-ICCV2023} as a recent GAN-based model for high-resolution face generation. Beyond GANs, we include SD3.5, an open-weight text-to-image diffusion model capable of synthesizing high-resolution face images at scale~\cite{SD3.5-news, SD3.5-large}. We additionally include GPT-5~\cite{GPT-5}, OpenAI's image generation model accessed via the GPT-5 interface after its August 2025 release, to represent a widely deployed closed-weight model baseline. Table~\ref{table:deepfake_models} summarizes the generators used for our work.

\subsection{Automated Deepfake Detection}

Automated detection has been extensively studied as a technical response to AI-generated images and broader synthetic media~\cite{Xu:AdvancementsAIGC:ACMCS2025, Wang:DeepfakeDetectionSurvey2025}. Most approaches train deep models to distinguish real images from generated ones by learning statistical or visual discrepancies from labeled datasets. Within this broader area, deepfake face detection remains one of the most mature and widely studied branches~\cite{Guo:ldfnet2023}. Existing methods have examined visible synthesis and blending artifacts~\cite{Li-exposing-arxiv2018, Li-FaceXray-CVPR2020}, frequency-domain inconsistencies~\cite{Qian-F3Net-ECCV2020}, and subtle trace amplification~\cite{Zhao-Mat-CVPR2021}. Under constrained experimental settings, some detection methods achieved very high accuracy~\cite{Ba-UMFC-AAAI2024}.

Nevertheless, detection performance often degrades under realistic conditions due to distribution shifts~\cite{Zi:WildDeepfake2020}, such as unseen deepfake generation models, model updates, image compression, or post-processing. In addition, automated detectors are not always available or accessible for everyday use. Even when such tools are available, users still need to interpret detector outputs and decide how to act. These limitations motivate complementary human-centered support for deepfake detection. Accordingly, our work focuses on identifying and organizing human-recognizable visual artifacts produced by deepfake generators, and on developing a taxonomy that supports humans in detecting deepfake faces.

\subsection{Human Detection of Deepfakes}

Empirical studies on human detection of deepfake content remain limited relative to automated detection. Existing work has examined synthetic media across modalities, including image, audio, text, and video~\cite{Alexander-human-CHBR2024, Pu:DeepfakeTextDetection:SP2023, Frank-DeepfakeHumanDetection-SP2024}. In the image domain, prior studies have considered both face images and broader image categories such as landscapes, animals, food, and objects~\cite{Lu:seeing2023, Cooke:Asgood2025}. Focusing on deepfake face images, we identified nine studies that reported quantitative human accuracy for photorealistic facial deepfakes~\cite{Bray-testinghuman-JC2023, Nightingale-Farid:PNAS2022, Boyd:value2023, Hulzebosch:DetectingCNN-generated2020, Diel:Inabilitytodetect2024, Shen:study2021, Frank-DeepfakeHumanDetection-SP2024, Lu:seeing2023, Cooke:Asgood2025} (see Table~\ref{tab:related_works} in Appendix~\ref{app:related_work}). Despite differences in experimental design, these studies converge on the following two conclusions.

\textbf{Unaided human performance is poor.}
Many past studies revealed that untrained participants tended to operate at a near-chance level on detecting high-quality deepfake faces~\cite{Nightingale-Farid:PNAS2022, Diel:Inabilitytodetect2024, Shen:study2021, Frank-DeepfakeHumanDetection-SP2024, Cooke:Asgood2025}, indicating limited spontaneous sensitivity to contemporary synthesis artifacts. For instance, Shen et al.~\cite{Shen:study2021} characterized baseline performance on StyleGAN2-generated faces as indistinguishable from random guessing. This perceptual limitation extends to large-scale settings involving image-face conditions, where Frank et al.~\cite{Frank-DeepfakeHumanDetection-SP2024} demonstrated that poor detection is a systemic failure rather than an artifact of specific participant pools or experimental designs. Such results validate the central premise that end users, despite being the primary targets of synthetic media, cannot reliably distinguish high-quality deepfakes without support.

\textbf{Interventions may help improve human performance.}
It has been reported that interventions based on training or feedback can improve human performance~\cite{Nightingale-Farid:PNAS2022, Boyd:value2023, Hulzebosch:DetectingCNN-generated2020, Diel:Inabilitytodetect2024}, but the improvements are modest, unstable across settings and studies, and strongly contingent on the specific training protocols and evaluation conditions employed. Feedback-based protocols serve as a primary mechanism for this improvement: Hulzebosch et al.~\cite{Hulzebosch:DetectingCNN-generated2020} demonstrated that iterative feedback refines accuracy on GAN-generated faces, while Diel et al.~\cite{Diel:Inabilitytodetect2024} found these gains robust when feedback is paired with specific artifact attention. Nightingale et al.~\cite{Nightingale-Farid:PNAS2022} further substantiated the utility of cue-based guidance, showing that artifact-focused tutorials on StyleGAN2 synthesis yield measurable discriminative improvements. Distinct from cognitive training, Boyd et al.~\cite{Boyd:value2023} validated the integration of AI support systems, confirming that external algorithmic assistance provides a distinct pathway to augmented accuracy on deepfake image. Despite the positive evidence reported, the effectiveness of interventions remains inconsistent across studies. For instance, Bray et al.~\cite{Bray-testinghuman-JC2023} observed negligible performance gains despite the provision of explicit knowledge regarding visible rendering artifacts, a result that decouples information availability from practical learnability. This finding implies that abstract awareness of artifacts does not automatically resolve into reliable perceptual heuristics. The literature thus characterizes training and assistance as promising yet volatile mechanisms, necessitating more systematic and actionable guidance to overcome the substantial variability observed across existing protocols.

Our analysis of the literature led to the discovery of some critical gaps. First, most studies focused on GAN-based models, especially StyleGAN2. Although some work studied multiple models including GAN-based models and diffusion models, model-specific effects have not been studied systematically. Therefore, if and how human performance and training effects vary across models remain unclear. Second, prior interventions often used artifacts as training cues, but these cues were selected without a systematic face-specific taxonomy. Consequently, the training and feedback interventions tested in prior work are less systematic.

Existing taxonomy-oriented work shows that organizing visual artifacts can support deepfake detection and related training materials. Kamali et al.~\cite{Negar:AI-Generated:2024} categorized observable cues in AI-generated photographs into anatomical, stylistic, functional, physical, and sociocultural categories. Their follow-up work further characterized artifacts in diffusion-generated images through a large-scale human study~\cite{Kamali:Photorealism:CHI2025}. These efforts show the value of structured artifact descriptions, but their scope mainly focuses on general AI-generated photographs and diffusion-model outputs. Their work does not cover model-specific and more fine-grained artifacts in deepfake face images.

Accordingly, we selected five representative generative models from three model families (GAN-based models, diffusion models, and MLLMs; see Table~\ref{table:deepfake_models}). We systematically characterize human-recognizable visual artifacts in fully generative face images and organize them into the Deepfake Face Artifact Taxonomy (DFAT), a training-oriented taxonomy for comparing artifacts across model families. DFAT provides a structured vocabulary of visual artifacts that can guide the construction of training materials and enable consistent analysis across models and participants.

\section{Deepfake Face Artifact Taxonomy}
\label{sec:DFAT}

\begin{figure*}[!tb] 
\centering
\includegraphics[width=\linewidth]{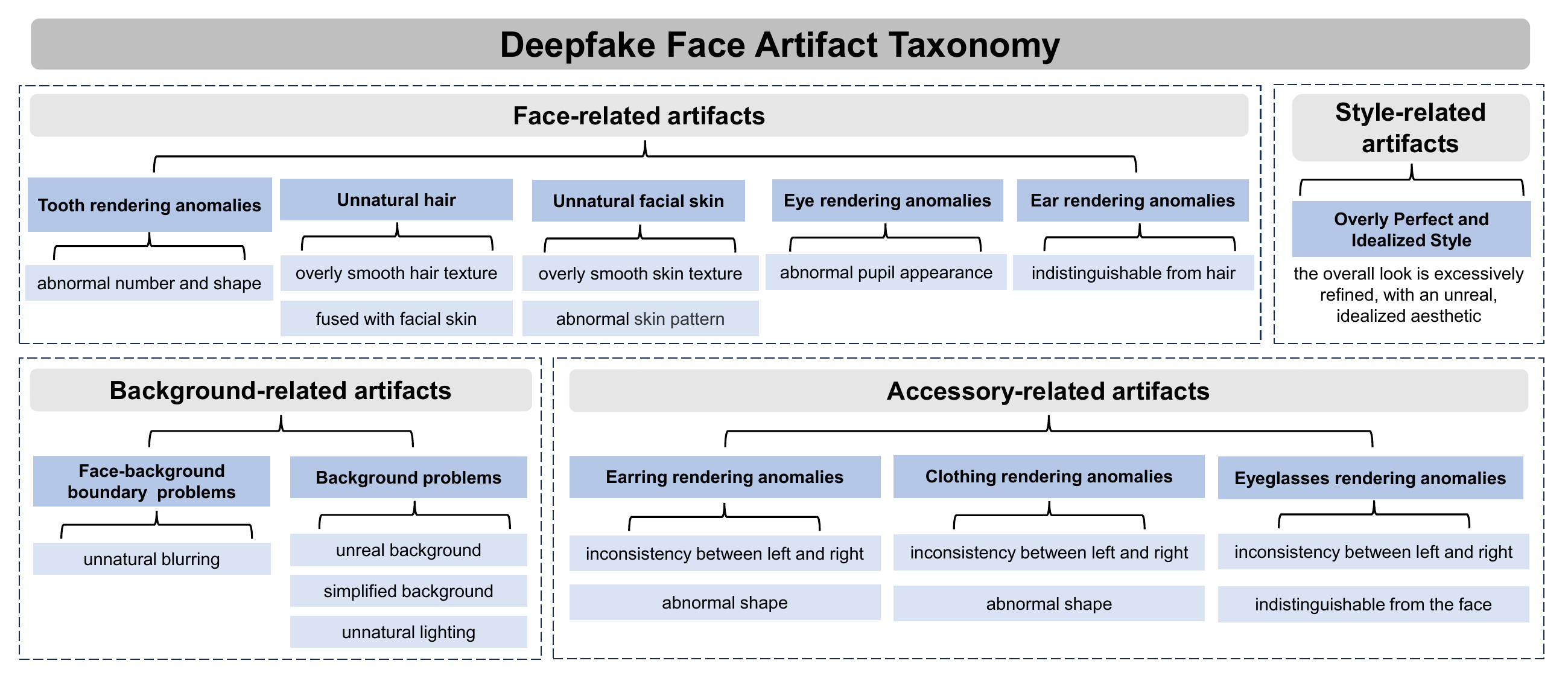} 
\caption{Illustration of the Deepfake Face Artifact Taxonomy (DFAT).}
\label{fig:Taxonomy}
\end{figure*}

This section describes our Deepfake Face Artifact Taxonomy (DFAT) with greater details, which addresses RQ1.

\subsection{Taxonomy Construction Methodology}

DFAT focuses on fully generative, photorealistic face images. We constructed DFAT by analyzing 1,000 $1024 \times 1024$ images, including 500 real photographs and 500 deepfake images generated by the five models shown in Table~\ref{table:deepfake_models}. Each deepfake model contributed 100 images. Real images were sourced from the Flickr-Faces-HQ (FFHQ) dataset~\cite{Karras-FFHQ-CVPR2019} to provide high-resolution reference images for comparison with synthetic outputs. To generate deepfake images, we used model-specific procedures:
\begingroup
\setlength{\IEEEilabelindent}{0pt}
\begin{itemize}
\item \textbf{GAN-based models (StyleGAN2, StyleGAN3, GRAM-HD)}: We used the pre-trained weights provided by the original authors. This setup reflects a common use case in which users directly deploy off-the-shelf generators to generate deepfake images.

\item \textbf{One diffusion model (SD3.5)}: We used SD3.5 through a local deployment to generate photorealistic face images via text-to-image synthesis. The prompts described portraits and varied subject attributes and image details to improve realism and cover diverse facial appearances.

\item \textbf{One MLLM (GPT-5)}: We used GPT-5 via the ChatGPT web interface to generate photorealistic face images. The prompts followed the same strategy, varying subject attributes and image details to improve realism and cover diverse facial appearances.
\end{itemize}
\endgroup

Model source links and all prompts are provided in the artifact (Appendix~\ref{app:artifact_availability}). We used direct outputs from the selected models as the deepfake images for DFAT, without manual retouching, compositing, or region-level replacement. We then manually inspected the images to identify human-visible anomalies that could support visual detection. We first recorded recurring low-level observations, such as distorted eyes, blurred boundaries, or inconsistent accessories. We then grouped related observations into broader artifact categories and organized these categories into the three-level hierarchy shown in Figure~\ref{fig:Taxonomy}. The resulting taxonomy captures four artifact domains, their artifact categories, and concrete visual manifestations.

\begin{table*}[!t] 
\centering
\small
\caption{\textbf{Prevalence of artifact types across deepfake models.} We categorize the occurrence frequency into four levels: \textbf{High} ($>$50\%), \textbf{Medium} ($>20\%$ and $\leq50\%$), \textbf{Low} (1--20\%), and \textbf{--} (not observed, i.e., 0\%).}
\label{tab:artifacts}
\begin{minipage}{0.9\textwidth} 
\centering
\renewcommand{\thefootnote}{\fnsymbol{footnote}}
\begin{tabular}{c ccccc}
\toprule
\textbf{Artifact Type} & \textbf{StyleGAN2} & \textbf{StyleGAN3} & \textbf{GRAM-HD} & \textbf{SD3.5} & \textbf{GPT-5}\\
\midrule
Tooth rendering anomalies (abnormal number/shape) & Low & Medium & Medium & -- & --\\
Unnatural hair (overly smooth texture) & -- & -- & Low & Medium & Medium\\ 
Unnatural hair (fused with facial skin) & Medium & High & Medium & Medium & Low\\
Unnatural facial skin (overly smooth texture) & -- & -- & -- & High & Low\\
Unnatural facial skin (abnormal skin pattern) & Low & High & High & High & --\\
Eye rendering anomalies (abnormal pupil appearance) & -- & -- & Low & Low & -- \\
Ear rendering anomalies (indistinguishable from hair) & Low & Medium & Medium & Low & --\\

\midrule
Overly perfect and idealized style & -- & -- & -- & High & High\\

\midrule
Face-background boundary (unnatural blurring) & Medium & Medium & High & High & --\\
Background problems (unreal background) & Medium & Medium & Medium & Low & --\\
Background problems (simplified background) & Medium & Medium & Low & High & Medium\\
Background problems (unnatural lighting) & -- & -- & Low & Low & -- \\

\midrule
Earring anomalies (left/right inconsistency)$^*$ & High & High & High & Medium & --\\
Earring anomalies (abnormal shape)$^*$ & Medium & Medium & High & Low & --\\
Clothing anomalies (left/right inconsistency) & Medium & Medium & Low & Low & --\\
Clothing anomalies (abnormal shape) & Medium & Medium & Medium & Low & --\\
Eyeglasses anomalies (left/right inconsistency)$^*$ & High & High & High & -- & --\\
Eyeglasses anomalies (indistinguishable from face)$^*$ & Medium & Medium & Medium & -- & --\\
\bottomrule
\multicolumn{6}{p{\linewidth}}{\footnotesize$^*$For accessory-related artifacts, the level indicates the proportion of images exhibiting at least one such artifact out of all images with one or more relevant accessories (not out of all images).}
\end{tabular}
\end{minipage}
\end{table*}

\subsection{Taxonomy Overview}
\label{subsec:DFAT_overview}

DFAT organizes visual artifacts into four primary domains: \textbf{face-related}, \textbf{style-related}, \textbf{background-related}, and \textbf{accessory-related}. Each domain is divided into \textit{artifact categories} (e.g., tooth rendering anomalies) and concrete \textit{visual manifestations} (e.g., abnormal number or shape). Five annotators each labeled 100 images from one model, recording the artifact types present. One co-author reviewed all annotations and finalized any corrections through discussion with each annotator.
Table~\ref{tab:artifacts} shows the prevalence levels of each artifact. These levels summarize the model-specific artifact profile observed in our sampled images and motivate the artifact cues used in the training materials.

\subsubsection{Face-related Artifacts}

Since the face is the primary focus of most models, these artifacts are often subtle and require careful observation across various facial features. We identified the following five categories of such artifacts.

\noindent\textbf{Tooth rendering anomalies.} While teeth may appear regular at first glance, models often struggle with the unique and complex shapes of individual teeth. This results in \textit{abnormal number and shape}, such as three prominent front teeth or malformed dental arrangements, highlighting the difficulty in processing multiple similar anatomical structures.

\noindent\textbf{Unnatural hair.} Accurately rendering the complex structure and intricate illumination of thousands of independent hairs remains a challenge. This is manifested in \textit{overly smooth hair textures} that lack depth and strand-level detail (e.g., common in SD3.5-generated images), and \textit{hair fused with facial skin}, where strands unnaturally merge with the skin or exhibit anomalous clumping and orientations.

\noindent\textbf{Unnatural facial skin.} Synthesizing realistic skin with natural light response is a complex task. Artifacts include \textit{overly smooth skin texture}, where the result lacks high-frequency details like pores (prevalent in SD3.5-generated images and occasionally observed in GPT-5-generated images), and \textit{abnormal skin patterns}, where the texture resembles a painted surface or exhibits subtle architecture-related inconsistencies. The latter is surprisingly noticeable in StyleGAN3, where we identified a previously unreported artifact: the skin often contains strange oval/diamond-shaped local mini-textures. Such patterns are difficult to see at normal viewing size but become clearly recognizable after magnification (see Appendix~\ref{app:stylegan3_examples} for examples).

\noindent\textbf{Eye rendering anomalies.} They primarily manifest as \textit{unnatural pupil appearances}, where pupils may be irregular in shape, inconsistent in size between eyes, or exhibit ghosting.

\noindent\textbf{Ear rendering anomalies.} They typically involve the ear being \textit{indistinguishable from hair}. For images with such artifacts, the model failed to separate contours, causing the ear to appear blurred or fused with adjacent hair strands.

\subsubsection{Style-related artifacts}

Unlike localized glitches, style-related artifacts pertain to the overall aesthetic composition and stylistic signature. Diffusion models and MLLMs like SD3.5 and GPT-5 are often over-optimized so that the generated images have a look and feel of being \textbf{overly perfect and idealized}. This results in a distinct, often unrealistic visual signature characterized by hyper-realism and ``idealized'' outputs that deviate from the natural variations found in genuine photography.

\subsubsection{Background-related Artifacts}

Background anomalies typically arise from generative designs that prioritize facial regions over peripheral content. Furthermore, training datasets often contain diverse and unstructured background information, making it challenging for models to learn semantically consistent representations of the background. There are two categories.

\noindent\textbf{Face-background boundary problems.} These occur at the interface between the subject (face, hair or shoulders) and the background. The primary manifestation is \textit{unnatural blurring}, where the face appears excessively sharp relative to the environment, or hair edges exhibit uneven blurring that is inconsistent with realistic depth-of-field effects.

\noindent\textbf{Background problems.} This category encompasses three distinct manifestations: (1) \textit{unreal background}, where the model generates meaningless or physically impossible textures; (2) \textit{simplified background}, where models (notably SD3.5) produce overly simplistic or blurry regions that lack the intricate details of real environments; and (3) \textit{unnatural lighting}, characterized by illumination inconsistencies within the background or between the subject and the scene, such as illogical light sources or absent shadows.

\subsubsection{Accessory-related Artifacts}

Deepfake models often have difficulties with rendering accessories to maintain semantic coherence, therefore leading to noticeable inconsistency issues that serve as crucial identification clues. There are three categories we observed.

\noindent\textbf{Earring rendering anomalies.} Earrings are small but highly diverse in design and material, often featuring complex reflective properties. Anomalies include \textit{inconsistency between left and right} (differences in size, shape, or color) and \textit{abnormal shapes} that do not conform to realistic designs or appear unnaturally attached to the earlobe.

\noindent\textbf{Clothing rendering anomalies.} Due to the immense variety in fabric materials and textures, clothing often exhibits \textit{left-right inconsistency} in symmetrical parts like collars or shoulder patterns. Additionally, \textit{abnormal shapes} may occur, such as impossible seams or textures that are inconsistent with known fabrics.

\noindent\textbf{Eyeglasses rendering anomalies.} As precisely structured objects, eyeglasses frequently exhibit \textit{left-right inconsistency} in frame geometry and lens size. In some cases, they are \textit{indistinguishable from the face}, where the frames unnaturally merge with the facial skin, indicating a failure to enforce clear object boundaries.

\subsection{From DFAT to Training Materials}
\label{subsec:development_training_materials}

The taxonomy served as the basis for designing our training materials that were used to train participants where to look and what visual cues to inspect for deepfake images. We selected artifacts using three criteria: frequency in the sampled images, visual salience, and human recognizability. These criteria prioritized cues that could be shown visually and practiced consistently, including background consistency, overall style, facial texture, and accessory symmetry. Prior research has shown that targeted guidance can improve deepfake detection performance~\cite{Boyd:value2023, Hulzebosch:DetectingCNN-generated2020, Diel:Inabilitytodetect2024}. Building on this evidence, our training materials use DFAT as a structured vocabulary for teaching human-visible artifacts across different deepfake models.

\subsubsection{Selecting Learnable Artifact Cues}

DFAT showed that different deepfake models share several broad artifact domains, such as background consistency, overall style, facial texture, and accessory symmetry. We used these shared domains to design general-purpose training material that teaches participants a common vocabulary and inspection routine across models. At the same time, the manifestations of these artifacts vary by model and differ in how easily they can be recognized. For example, SD3.5 and GPT-5 images often exhibit global cues such as an overly polished or idealized visual style, while StyleGAN3 images may contain abnormal skin-texture patterns that become recognizable only under magnification. We therefore designed separate model-specific training materials to focus on the most recognizable cues associated with each deepfake model.

\subsubsection{Constructing the Training Materials}

We designed two sets of short instructional materials: one general-purpose set covering all five selected deepfake models, and one model-specific set for each model. The general-purpose set includes a five-minute video and ten interactive web-based practice questions. It was designed to cover a broader range of artifact domains across models and teach participants a shared vocabulary and inspection routine for deepfake faces. Each model-specific set includes a one- to three-minute video and five interactive practice questions focused on artifacts associated with the corresponding model, based on its artifact profile shown in Table~\ref{tab:artifacts}. Having these two sets allowed us to compare whether broad artifact awareness and model-specific guidance can have different effects on improving human performance on deepfake detection (RQ3).

For each practice question, the system displayed an explanation of the correct answer to help participants connect the visual cue to the detection judgment.
We chose short videos as the training format because they have been found effective for educating on security-related topics~\cite{Chen:Empowering:ACMCCS2025, kaushik2021know, berens2024taking}. The videos follow a global-to-local inspection routine: they first guide viewers to inspect image-level cues, such as the background and overall style, and then ask them to zoom in on localized cues around the face and accessories. This ordering was intended to help participants form a consistent inspection routine rather than memorize isolated features.

\section{Online Experiment}
\label{sec:online_experiment}

Building on the DFAT-guided training materials, we designed and conducted an online experiment to evaluate how human performance in detecting deepfakes and identifying subjective states varies across different models and training conditions. The main experiment addresses \textbf{RQ2} and \textbf{RQ3} by measuring detection performance for the five deepfake models before and after DFAT-guided training (general-purpose vs.~model-specific). To address \textbf{RQ4}, we invited participants back five months later and tested retention and transfer on newly sampled deepfake images they had not seen before. This section describes the study design, procedure, stimuli, participants, measures, and statistical analyses.

\subsection{Study Design}
\label{subsec:study_design}

We used a mixed design with two between-subject factors and two within-subject factors. The between-subject factors were training type (general-purpose vs.~model-specific) and deepfake model condition (StyleGAN2, StyleGAN3, GRAM-HD, SD3.5, or GPT-5). Participants were randomly assigned to one training group and one model condition. The within-subject factors were testing session (pre-training vs.~post-training) and presentation format (single-image vs.~side-by-side).

The two presentation formats captured different detection settings. In the \textbf{single-image} format, participants judged one face image in isolation, reflecting scenarios such as social media profile photos or shared photographs where users usually lack an authentic reference. In the \textbf{side-by-side} format, participants compared one real and one deepfake image and were asked to identify the real one. This format captures reference-assisted detection, such as educational quizzes or platform-assisted settings where a verified image is shown next to a suspicious one~\cite{WhichFaceIsReal}. Pairwise judgments are also widely used in subjective image-quality assessment~\cite{Zhang:Subjective2023, Mohammadi:Subjective2023, Li:Quality2021}. Each participant completed both formats before and after training, with ten questions per format in each testing session. The follow-up study retained each participant's original training group and generative model.

We measured detection performance as the proportion of correct judgments in each task condition. For the single-image subtask, we also computed deepfake-image detection accuracy and real-image detection accuracy to examine whether training affected responses to deepfake and real images differently. After each testing session, participants reported task-specific confidence and stress levels on a five-point scale. These measures supported auxiliary analyses related to their subjective states before and after training.

\subsection{Participants}
\label{subsec:participants_exclusion}

Participants were recruited from \textit{Prolific}, an online platform widely used by researchers for human experiments. We restricted eligibility to adults who were fluent in English to ensure comprehension of the task description. Before recruiting participants, we conducted a power analysis to calculate the target sample size for the planned mixed design. Using G*Power with detection accuracy as the primary outcome, the analysis assumed a small effect ($f=.10$), $\alpha=.05$, power ($1-\beta$) $=.95$, 10 between-subject groups, and four repeated measurements, yielding a target sample size of 440 participants. Participants were informed about the purpose, structure, and approximate duration of the study before providing informed consent through the experiment website. To ensure data quality, we excluded participants who failed attention checks, had missing responses, or completed the tests in less than 6 minutes (see Appendix~\ref{app:data_control}).

A total of 608 \textit{Prolific} users participated in the experiment. After applying the exclusion criteria, 524 valid participants remained for analysis: 278 in the general-purpose training group and 246 in the model-specific group, distributed across the five model conditions. On average, it took each participant 32.3 minutes (general-purpose) and 29.5 minutes (model-specific) to complete the main study. Based on demographic profiles, participants were categorized by gender (Male, Female), ethnicity (White, Black, Asian), and age group (Table~\ref{tab:demographics} in Appendix~\ref{app:data_control} reports more details).

Five months later, 342 participants from the original valid sample (181 from the general-purpose group and 161 from the model-specific one) completed the follow-up study.

\subsection{Procedure}
\label{subsec:procedure}

\subsubsection{Main Study}

The main study had four phases: consent and demographics, pre-training test, DFAT-guided training with practice questions, and post-training test.

\noindent\textbf{Consent and demographics.} After reading the consent form, participants could either proceed or leave the study. They reported demographic information, including age, gender, and ethnicity. Before the first test, the system explained the task interface and response requirements.

\noindent\textbf{Pre-training test.} Participants completed 20 baseline questions before receiving any training:
\begingroup
\setlength{\IEEEilabelindent}{0pt}
\begin{itemize}
\item \textbf{Single-image} (10 questions): Participants viewed one image and judged whether it was a real photograph or a deepfake.
\item \textbf{Side-by-side} (10 questions): Participants viewed two images, one real and one deepfake, and identified which image was real.
\end{itemize}
\endgroup
After the questions, participants reported their confidence and perceived stress using the self-assessment questions listed in Appendix~\ref{app:self_assessment}.

\noindent\textbf{Training.} Participants received either general-purpose or model-specific DFAT-guided training according to their assigned group. The training included a video and interactive practice questions. For each practice question, the system explained the reasons for the correct answer.

\noindent\textbf{Post-training test.} The post-training session used the same task structure as the pre-training session, with newly sampled images to reduce memory effects. Participants reported their confidence and perceived stress after completing the test. They also provided feedback on the training regarding perceived usefulness, memorable content, and open-ended comments (see Appendix~\ref{app:main_feedback_questions} for the list of questions).

\subsubsection{Follow-up Study}

Five months after the main study, we invited all participants in the final sample to complete a follow-up study. Without viewing the training materials again, participants repeated the post-training test with exactly the same materials as last time, including both single-image and side-by-side tasks. They then completed 20 additional single-image questions with newly sampled real and deepfake images. These additional trials tested whether previously learned cues could be transferred to unseen images. Subsequently, participants reported their confidence and perceived stress and completed a feedback questionnaire about visual cues, memorable training content, and open-ended comments (see Appendix~\ref{app:followup_feedback_questions}).

\subsection{Test Stimuli}
\label{subsec:deepfake_image}

For the main study, each participant completed a total of 40 questions across the pre- and post-training tests. The stimulus set consisted of 30 deepfake images created by the assigned model and 30 real photographs sourced from the FFHQ dataset~\cite{Karras-FFHQ-CVPR2019}. All stimuli were $1024 \times 1024$ images. Participants assigned to the same model were tested on the same image set. Across participants, we randomized the presentation order and interleaved the two subtasks (single-image and side-by-side) to reduce order effects and task-mode biases. The pre-training and post-training tests used non-overlapping stimuli, so that participants did not see the same test image twice across sessions.

For the follow-up study, the retention test used the same task structure and image set as the post-training test in the main study. The additional single-image trials used newly sampled real and deepfake images from the corresponding model condition to assess transfer to unseen images.

\subsection{Statistical Analyses}
\label{subsec:statistical_analysis}

Before examining training effectiveness, we confirmed that the two training groups did not differ in demographic distribution (Gender: $\chi^2(1) < 0.01$, $p = .979$; Ethnicity: $\chi^2(2) = 1.30$, $p = .521$; Age: $\chi^2(2) = 4.84$, $p = .089$). Because accuracy scores were non-normally distributed (Shapiro--Wilk, $p < .05$), we used non-parametric analyses: Aligned Rank Transform (ART) ANOVA~\cite{Wobbrock:ARTANOVA:CHI2011} for factorial designs and Wilcoxon signed-rank tests~\cite{Woolson:Wilcoxontest:2007Wiley} for paired comparisons. We report partial $\eta_p^2$ and rank-biserial correlation ($r_{\mathrm{rb}}$) as effect sizes, respectively. Unless stated otherwise, we apply Holm correction within each family of related post hoc or paired comparisons reported in Section~\ref{sec:results}.

\section{Results}
\label{sec:results}

Building on the DFAT that addresses \textbf{RQ1}, this section reports results for \textbf{RQ2--RQ4}: model-dependent detectability, training effectiveness, and five-month retention of detection and subjective-state outcomes. Table~\ref{tab:descriptive_stats_all} summarizes descriptive statistics of pre- and post-training detection accuracy in different conditions. We also report participants' feedback that contextualizes how participants perceived, used, and remembered the training cues.

\begin{table}[!thb]
\centering
\caption{Descriptive statistics for detection accuracy.}
\label{tab:descriptive_stats_all}
\resizebox{\columnwidth}{!}{%
\begin{tabular}{l l l c c c}
\toprule
\multirow{2}{*}{\makecell[c]{\textbf{Training}\\\textbf{materials}}} 
& \multirow{2}{*}{\textbf{Format}} 
& \multirow{2}{*}{\makecell[c]{\textbf{Deepfake}\\\textbf{model}}}
& \multicolumn{2}{c}{\textbf{Detection accuracy: mean (SD)}}
& \multirow{2}{*}{\textbf{N}} \\
\cmidrule(lr){4-5}
& & & \textbf{Pre-Training} & \textbf{Post-Training} & \\
\midrule
\multirow{10}{*}{\makecell[c]{General-\\purpose}} & \multirow{5}{*}{Side-by-side}
    & StyleGAN2 & 62.0\% (19.8\%) & 74.1\% (18.4\%) & 54 \\
& & StyleGAN3 & 67.5\% (25.8\%) & 81.6\% (20.9\%) & 55 \\
& & GRAM-HD   & 79.2\% (22.4\%) & 86.8\% (15.9\%) & 60 \\
& & SD3.5     & 92.7\% (12.9\%) & 93.2\% (12.1\%) & 60 \\
& & GPT-5     & 80.8\% (21.2\%) & 85.3\% (18.7\%) & 49 \\
\cmidrule(l){2-6}
& \multirow{5}{*}{Single image}
    & StyleGAN2 & 56.5\% (22.8\%) & 67.6\% (17.6\%) & 54 \\
& & StyleGAN3 & 65.3\% (21.2\%) & 72.0\% (19.2\%) & 55 \\
& & GRAM-HD   & 72.8\% (18.7\%) & 77.3\% (16.6\%) & 60 \\
& & SD3.5     & 88.0\% (15.5\%) & 81.2\% (12.8\%) & 60 \\
& & GPT-5     & 75.1\% (22.5\%) & 75.7\% (18.1\%) & 49 \\
\midrule
\multirow{10}{*}{\makecell[c]{Model-\\specific}} & \multirow{5}{*}{Side-by-side}
    & StyleGAN2 & 62.0\% (21.4\%) & 74.6\% (17.1\%) & 50 \\
& & StyleGAN3 & 66.5\% (25.6\%) & 89.2\% (13.4\%) & 51 \\
& & GRAM-HD   & 77.8\% (18.6\%) & 93.7\% (9.5\%) & 49 \\
& & SD3.5     & 97.2\% (5.7\%)  & 98.8\% (3.3\%)  & 50 \\
& & GPT-5     & 89.1\% (18.0\%) & 96.1\% (10.4\%) & 46 \\
\cmidrule(l){2-6}
& \multirow{5}{*}{Single image}
    & StyleGAN2 & 58.0\% (17.5\%) & 68.4\% (18.1\%) & 50 \\
& & StyleGAN3 & 65.1\% (23.6\%) & 79.2\% (15.7\%) & 51 \\
& & GRAM-HD   & 70.0\% (19.7\%) & 83.3\% (14.9\%) & 49 \\
& & SD3.5     & 89.4\% (11.9\%) & 97.4\% (6.3\%)  & 50 \\
& & GPT-5     & 82.0\% (18.6\%) & 93.9\% (8.8\%)  & 46 \\
\bottomrule
\end{tabular}%
}
\end{table}

\subsection{Baseline Detection Performance}
\label{subsec:detection_performance}

\subsubsection{Baseline Detectability Is Model-Dependent}

We use pre-training accuracy to assess \textbf{RQ2} because it captures unaided detection performance before training materials could have affected participants' judgments. A two-way ART ANOVA on pre-training accuracy showed a significant main effect of deepfake model ($F(4, 519) = 64.02$, $p < .001$, $\eta_p^2 = 0.33$). The image presentation format also had a significant main effect ($F(1, 519) = 35.21$, $p < .001$, $\eta_p^2 = 0.064$), with the side-by-side setting yielding higher overall accuracy. The deepfake model $\times$ format interaction was not significant ($F(4, 519) = 1.12$, $p = .347$, $\eta_p^2 = 0.009$), suggesting that the relative detectability of the models was broadly consistent across presentation formats.

\begin{figure}[!tbp]
\centering
\includegraphics[width=\linewidth]{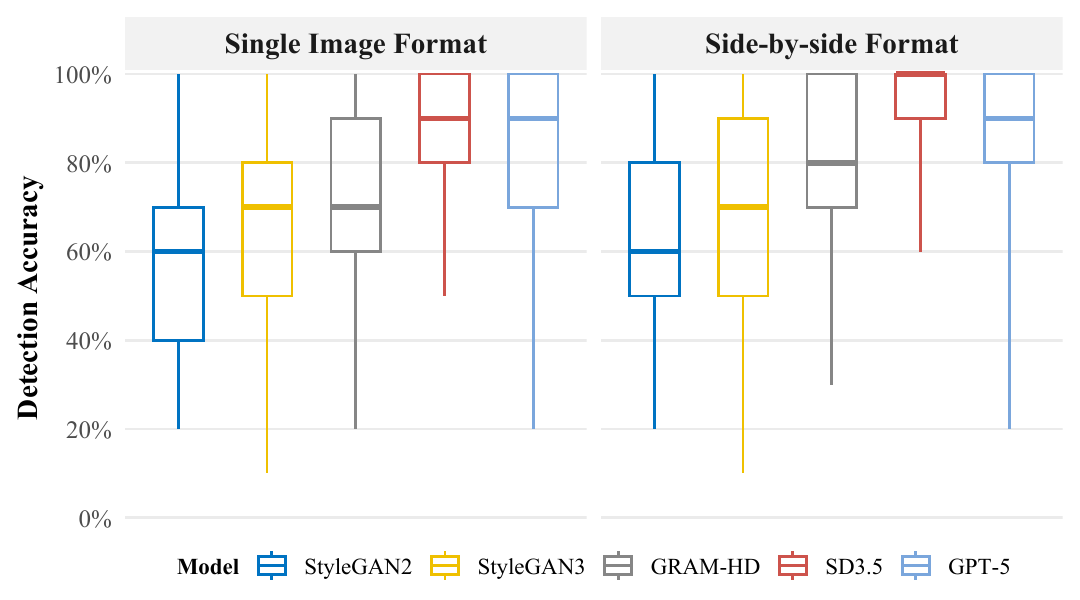}
\caption{Baseline detection accuracy across deepfake models and presentation formats. Box plots display median, interquartile range, and full range.}
\label{fig:pretest_by_model}
\end{figure}

Figure~\ref{fig:pretest_by_model} shows the baseline patterns of detection performances for the five models: images generated by SD3.5 and GPT-5 were easier to detect, images generated by GRAM-HD fell in the middle, and images generated by StyleGAN2 and StyleGAN3 were more difficult. Holm-corrected post hoc comparisons supported this broad ordering across presentation formats. The adjacent differences among the upper three groups were reliable in both formats: SD3.5 had higher baseline accuracy than GPT-5 ($p_{\mathrm{adj}}\le .0019$), GPT-5 had higher baseline accuracy than GRAM-HD ($p_{\mathrm{adj}}\le .032$), and GRAM-HD had higher baseline accuracy than StyleGAN2 ($p_{\mathrm{adj}}<.001$). The boundary between GRAM-HD and StyleGAN3 depended on presentation format: the difference was reliable in the side-by-side format ($p_{\mathrm{adj}}=.005$), but did not survive correction in the single-image format ($p_{\mathrm{adj}}=.074$). These results align with predictions based on DFAT: higher baseline accuracy for SD3.5 and GPT-5 is consistent with global cues, such as an overly polished or idealized visual style, being easier for untrained observers to notice; lower baseline accuracy for StyleGAN-based models was linked to texture-level and localized artifacts that are considered harder to identify without explicit guidance.

Two secondary analyses further bounded this baseline pattern. First, side-by-side references improved baseline accuracy for four of the five deepfake models. The format advantage was largest for SD3.5 ($p < .001$, $r_{\mathrm{rb}} = 0.74$) and was also significant for GPT-5, GRAM-HD, and StyleGAN2; StyleGAN3 was the only exception ($p = .398$, $r_{\mathrm{rb}} = 0.11$). Second, demographic effects were limited. Age was the only variable with a consistent main effect across both presentation formats (both $p$s $< .001$), with participants aged 18--24 achieving higher baseline accuracy than older groups, especially for GPT-5, GRAM-HD, and SD3.5. Gender and ethnicity showed no significant main effects (all $p$s $\ge .076$). Although we observed one deepfake model $\times$ gender interaction in the side-by-side format ($p < .001$, $\eta_p^2 = 0.04$), it was model-specific and did not indicate a consistent gender advantage.

\begin{tcolorbox}[hylist]
\textbf{Finding 1 (RQ2):} Baseline detection performance varied across deepfake models with distinctive human-recognizable artifacts. Deepfakes generated by more recent and advanced models were not necessarily harder to detect, e.g., participants detected SD3.5 and GPT-5 images more accurately than StyleGAN2 and StyleGAN3 ones, and StyleGAN2 images were harder than StyleGAN3 ones. Side-by-side references improved baseline accuracy, but model-dependent differences remained across presentation formats.
\end{tcolorbox}

\subsection{Training Effects on Detection Performance}
\label{subsec:training_effects}

DFAT-guided training improved detection in most conditions, but the size of the gain depended on the deepfake model and the training material. We quantified training gains as post-training accuracy minus pre-training accuracy for each participant and compared gains across deepfake models, training-material types, and presentation formats.

\subsubsection{Training Gains Are Uneven Across Models}

\begin{figure}[tbp]
\centering
\includegraphics[width=\columnwidth]{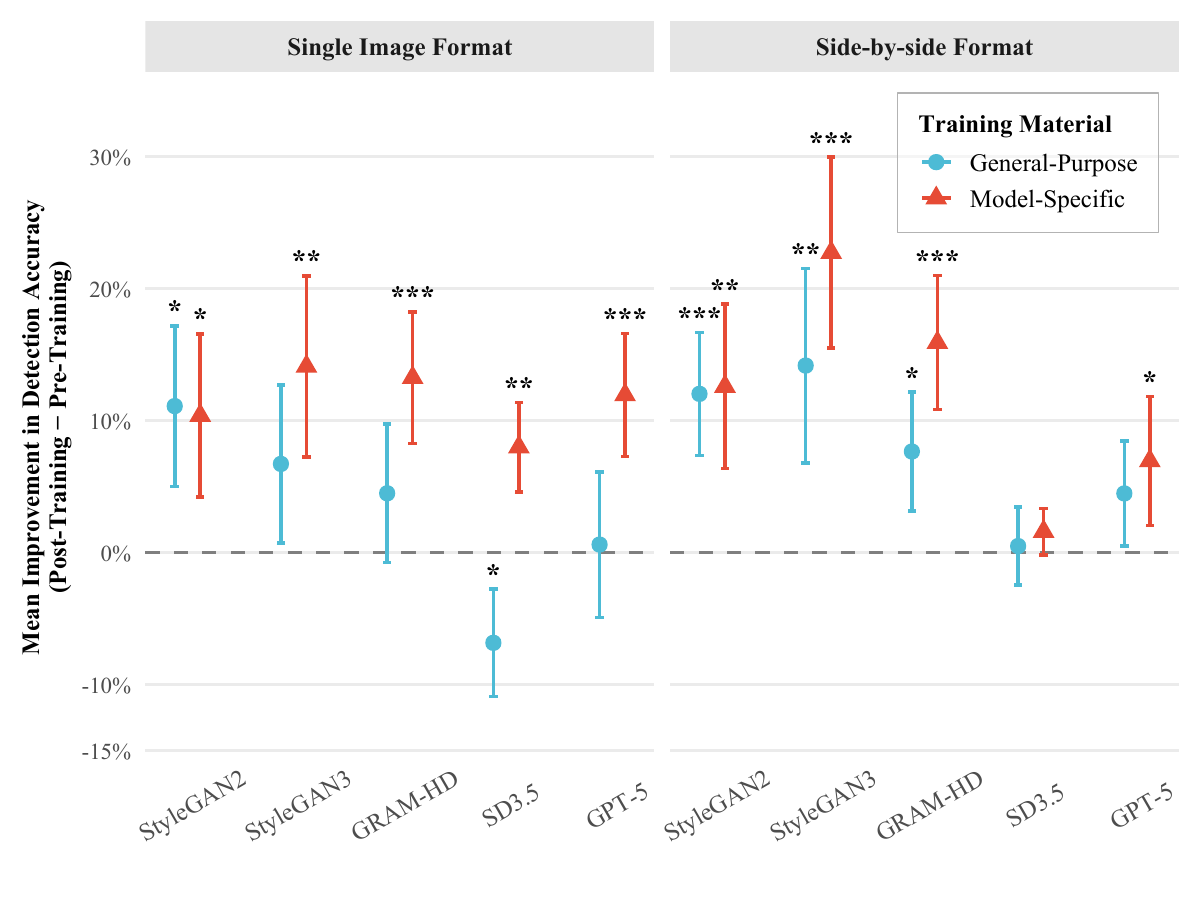}
\caption{Mean improvement in detection accuracy (post-training minus pre-training) by deepfake model, training material, and presentation format. Error bars show 95\% confidence intervals; significance markers denote Holm-corrected Wilcoxon signed-rank tests for pre--post changes.}
\label{fig:training_effect}
\end{figure}

Holm-corrected Wilcoxon signed-rank tests showed significant accuracy gains in 13 of the 20 experimental conditions. Figure~\ref{fig:training_effect} summarizes the model-level gain patterns: more consistent gains were observed for GAN-based models, whereas gains for SD3.5 and GPT-5 were more conditional. StyleGAN3 showed the largest improvement with model-specific training (side-by-side: \(+22.8\%\), \(p < .001\), $r_{\mathrm{rb}} = 0.87$). For SD3.5, baseline accuracy was already high ($\ge 88\%$), and significant gains appeared only in the single-image format with model-specific training ($+8.0\%$, $p = .001$, $r_{\mathrm{rb}} = 0.96$). General-purpose training was associated with a decrease for SD3.5 in the single-image condition ($-6.8\%$, $p = .015$, $r_{\mathrm{rb}}=-0.53$), consistent with general artifact cues being less well matched to SD3.5 images. For GPT-5, significant gains appeared with model-specific training, while general-purpose training produced no significant improvement.

Two-way ART ANOVAs showed that the gains depended on deepfake model ($p$s $< .05$). StyleGAN3 showed the largest average improvement (14.5\%). This improvement was similar to the gains for StyleGAN2 and GRAM-HD. It was larger than the gains for SD3.5 in most conditions ($p_{\mathrm{adj}}\le .016$), and larger than the GPT-5 gain only under model-specific training in the side-by-side format ($p_{\mathrm{adj}}=.0145$). These results suggest that participants benefited more consistently from training on GAN-related artifacts, whereas gains for SD3.5 and GPT-5 were more conditional, partly because their baseline accuracy was already high.

\subsubsection{Model-Specific Training Helped Most in Single-Image Subtasks}

Model-specific training provided the clearest advantage when participants could not rely on a side-by-side reference. Two-way ART ANOVAs comparing gains by training-material type showed significant training-material type $\times$ time (pre- and post-training) interactions in three of the ten model-format combinations: GRAM-HD in the single-image format ($F(1, 107) = 5.95$, $p = .016$, $\eta_p^2 = 0.053$), SD3.5 in the single-image format ($F(1,108)=65.74$, $p<.001$, $\eta_p^2=0.378$), and GPT-5 in the single-image format ($F(1,93)=8.03$, $p=.006$, $\eta_p^2=0.080$).

For GAN-based models, direct comparisons of performance gains between general-purpose and model-specific training were not statistically significant after Holm correction (all adjusted $p$ values $> .10$). Descriptively, model-specific training showed numerical advantages for StyleGAN3 (+8.6\% in side-by-side and +7.4\% in single-image) and GRAM-HD (+8.3\% in side-by-side and +8.8\% in single-image). These differences were not significant in direct gain comparisons, possibly reflecting limited sample sizes and high variance in participant performance.

The training-material difference was clearest for SD3.5 and GPT-5 in the single-image format. For SD3.5, model-specific training outperformed general-purpose training by 14.8\% ($p_{\mathrm{adj}} < .001$, $r_{\mathrm{rb}}=0.58$), with a 6.8\% performance decrease under general-purpose training and an 8.0\% increase under model-specific training. Similarly, GPT-5 benefited more from model-specific training in the single-image condition (difference in gains \(= +11.3\%\), $p_{\mathrm{adj}}=.022$, $r_{\mathrm{rb}}=0.35$), whereas general-purpose training produced almost no improvement (+0.6\%).

\subsubsection{Format and Demographic Effects Are Limited}

Presentation format did not substantially change the training-gain pattern. Two-way ART ANOVAs conducted separately for each deepfake model and training-material type showed significant Time $\times$ Format interactions in three of the ten conditions. Holm-corrected post hoc comparisons showed significant format-dependent gain differences only for SD3.5. Under general-purpose training, side-by-side presentation yielded larger gains than the single-image format ($p_{\mathrm{adj}}=.017$, $r_{\mathrm{rb}}=0.51$); under model-specific training, the pattern reversed, with larger gains in the single-image format ($p_{\mathrm{adj}}=.008$, $r_{\mathrm{rb}}=-0.81$). Gains were comparable across formats for all other deepfake models.

Demographic variables showed no consistent relationship with training gains. ART ANOVAs tested interactions between age, gender, ethnicity, and training gains across deepfake models, training material types, and presentation formats. Across 60 tests, only six interactions reached $p < .05$, and these effects did not form a consistent pattern across demographic variables or task conditions. Thus, although age predicted baseline accuracy, we found little evidence that training gains systematically differed across demographic groups. Such seemingly negative results deliver a positive signal: when training worked, it worked across different demographic groups.

\begin{tcolorbox}[hylist]
\textbf{Finding 2 (RQ3):} DFAT-guided training improved participants' detection accuracy in most conditions, and model-specific training yielded substantive advantage over general-purpose training in the single-image format. The size of the improvement varied across deepfake models and training types: GAN-based models showed more consistent gains, while SD3.5 and GPT-5 improved most clearly under model-specific training.
\end{tcolorbox}

\begin{figure*}[!t] 
\centering
\includegraphics[width=\linewidth]{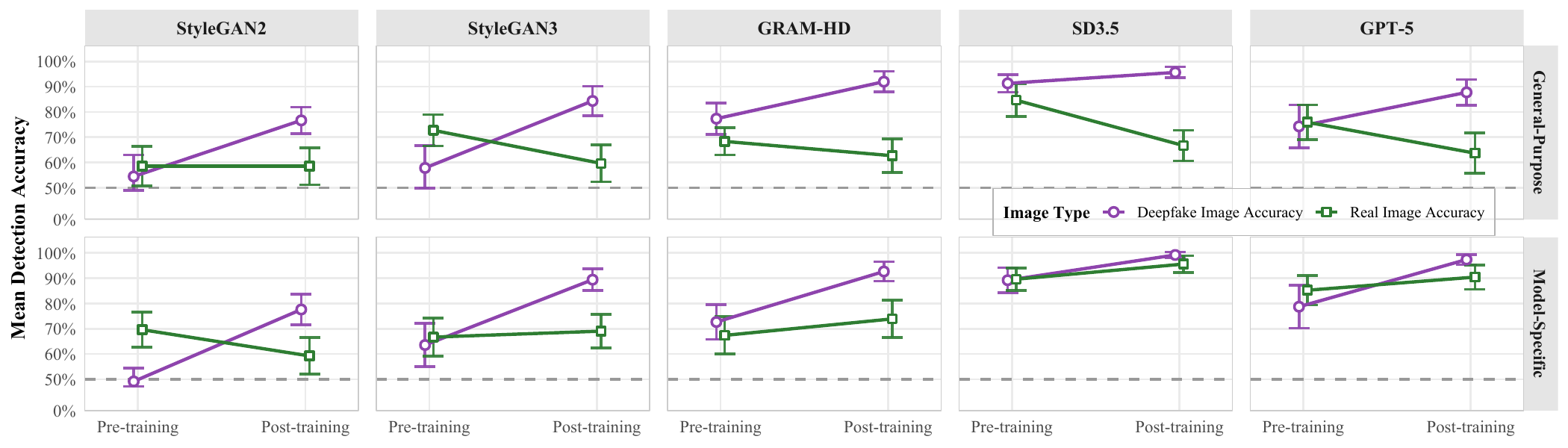} 
\caption{Pre--post changes in deepfake-image and real-image detection accuracy in the single-image task. Points show mean accuracy; error bars indicate 95\% confidence intervals.}
\label{fig:accuracy_patterns}
\end{figure*}

\subsection{Training Effects on Detection Behavior and Subjective States}
\label{subsec:behavior_subjective}

Aggregate training gains show whether overall accuracy improved, but they do not reveal whether gains came from better recognition of deepfake images, better recognition of real images, or both. To characterize how training changed single-image judgments, we decomposed accuracy by image type and analyzed confidence and perceived stress as subjective correlates of detection performance.

\subsubsection{Training Shifted Sensitivity Toward Deepfake Images}

DFAT-guided training improved deepfake-image accuracy more consistently than real-image accuracy. As shown in Figure~\ref{fig:accuracy_patterns}, deepfake-image accuracy increased in nine of the ten conditions after Holm correction. The only non-significant case was SD3.5 under general-purpose training, where baseline accuracy was already above 90\%. Real-image accuracy changed less consistently than deepfake-image accuracy. Under general-purpose training, real-image accuracy decreased for SD3.5 ($p_{\mathrm{adj}}<.001$) and for StyleGAN3 and GPT-5 (both $p_{\mathrm{adj}}=.048$). Under model-specific training, we observed no Holm-corrected decreases in real-image accuracy. Overall, training increased sensitivity to deepfake cues, and general-purpose training increased false positives on real images for a subset of deepfake models.

ART ANOVAs showed significant image type $\times$ time interactions for all five deepfake models under general-purpose training ($p$s $< .01$), but only for StyleGAN2 and StyleGAN3 under model-specific training ($p$s $< .001$). Post-training comparisons showed a consistent asymmetry: deepfake-image accuracy was higher than real-image accuracy in all conditions ($p$s $< .05$). These results suggest that training primarily increased participants' sensitivity to deepfake images, while real-image judgments remained stable except under general-purpose training.

\subsubsection{Confidence and Stress Level Change After Training}

We measured task-specific confidence and stress self-reported by participants after each testing session (5-point scales). In the main study ($n=524$), model-specific training increased confidence (pre-training: $M=3.41$, post-training: $M=3.78$; Holm-corrected $p_{\mathrm{adj}}<.001$). General-purpose training showed a smaller confidence change that did not survive Holm correction (pre-training: $M=3.49$, post-training: $M=3.57$; $p_{\mathrm{adj}}=.143$).

Self-reported stress level increased after general-purpose training (pre-training: $M=2.16$, post-training: $M=2.38$; $p_{\mathrm{adj}}<.001$). Model-specific training showed no Holm-corrected stress increase (pre-training: $M=2.09$, post-training: $M=2.00$; $p_{\mathrm{adj}}=.143$). As a whole, the results suggest that training can increase participants' confidence, while its effects on self-reported subjective states vary by training type. Considering the training can improve detection performance in most conditions, the overall effect of training is therefore positive in most cases or neutral in a few cases.

\begin{tcolorbox}[hylist]
\textbf{Finding 3 (RQ3):} DFAT-guided training effectively improved participants' performance in detecting deepfake images, with detection accuracy increased in nine out of ten single-image conditions. Under general-purpose training, real-image detection accuracy decreased for certain models, while this decrease was not found under model-specific training. Model-specific training increased confidence without increasing stress, whereas general-purpose training produced a small but significant stress increase.
\end{tcolorbox}

\subsection{Training Effects After Five Months}
\label{subsec:retention}

\begin{figure*}[!tb]
\centering
\includegraphics[width=\linewidth]{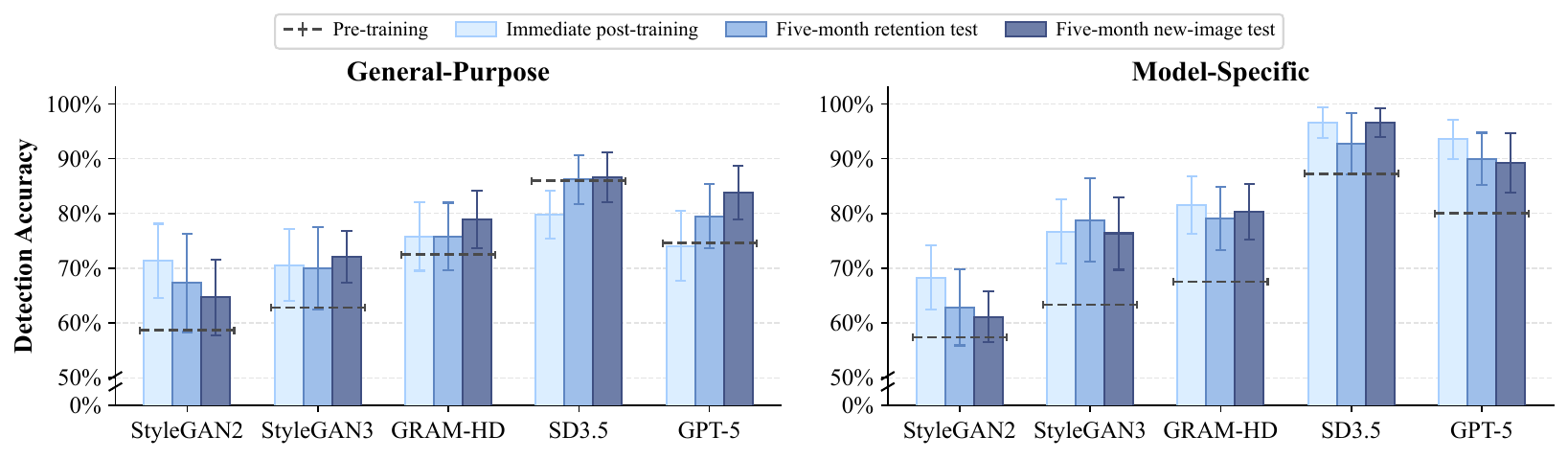}
\caption{Post-training detection accuracy across models and training material types: the main experiment (immediately after training), the follow-up test and the new-image test (five months later). Error bars indicate standard errors; annotations show changes from immediate post-training to five-month retention.}
\label{fig:retention_tests}
\end{figure*}

Five months after the main experiment, 342 participants completed the follow-up study. To answer \textbf{RQ4}, we compared immediate post-training and follow-up results for detection accuracy, newly sampled single-image trials, and self-reported confidence and stress levels.

\subsubsection{Detection Gains Showed Little Decay}

The retention test repeated the post-training test structure, allowing paired comparisons under the same deepfake-model condition and presentation format. As shown in Figure~\ref{fig:retention_tests}, follow-up accuracy generally remained close to immediate post-training accuracy and above the corresponding pre-training baseline.

Newly sampled single-image trials showed a similar descriptive pattern. Because these trials used real and deepfake images that were not matched to the stimuli from the main experiment, we interpret them as descriptive evidence of transfer rather than as part of the paired retention test.

For the formal retention analysis, we used paired Wilcoxon signed-rank tests to compare retention-test and immediate post-training accuracy for each combination of deepfake model, training-material type, and presentation format. Mean changes from post-training to follow-up were small, ranging from $-7.3\%$ to $+6.4\%$. After Holm correction, none of the 20 paired comparisons showed evidence of a significant decline in detection accuracy. The largest numerical decrease occurred for StyleGAN2 under general-purpose training in the side-by-side format ($-7.3\%$; post-training: $M = 76.3\%$, retention: $M = 69.0\%$), whereas the largest numerical increase occurred for SD3.5 under general-purpose training in the single-image format ($+6.4\%$; post-training: $M = 79.8\%$, retention: $M = 86.2\%$). Neither change was significant after correction.

\subsubsection{Confidence and Stress Levels Remained Stable}

We compared subjective-state ratings between the immediate post-training test and the five-month follow-up. Confidence showed no Holm-corrected decline in the general-purpose group (post-training: $M=3.49$, follow-up: $M=3.34$; $p_{\mathrm{adj}}=.071$). The model-specific group showed a modest decline (post-training: $M=3.67$, follow-up: $M=3.41$; $p_{\mathrm{adj}}=.001$). The self-reported stress level did not show a Holm-corrected increase in either training group (general-purpose: post-training $M=2.46$, follow-up $M=2.28$, $p_{\mathrm{adj}}=.089$; model-specific: post-training $M=2.11$, follow-up $M=2.20$, $p_{\mathrm{adj}}=1.000$). These results reconfirmed that applying learned cues after five months did not increase stress.

\begin{tcolorbox}[hylist]
\textbf{Finding 4 (RQ4):} The results show that the effects of DFAT-guided training persisted over five months. Detection accuracy did not decline significantly in any paired comparison, and the new image tests showed a similar pattern. Self-reported stress showed no increase at follow-up, while confidence exhibited limited decay for those who received model-specific training.
\end{tcolorbox}

\subsection{Participant Feedback}
\label{subsec:participant_feedback}

In total, 520 (out of 524) participants in the main study and 341 (out of 342) participants in the five-month follow-up provided feedback after completing the test. We analyzed these responses to understand how participants perceived the training, which visual artifacts they used after training, and which ones remained salient five months later.

\textbf{Most participants perceived the training as useful.}
In the main study, 89.9\% of the participants reported that the training improved their ability to identify deepfake face images. Participants valued both training components: 83.4\% considered the training video useful, 68.3\% considered the practice tasks useful, and 52.9\% selected both. Among the artifacts participants relied on during the tests, the most frequently reported were unnatural facial skin (71.9\%), background problems (69.7\%), accessory rendering anomalies (68.3\%), overly perfect and idealized style (66.2\%), and unnatural hair (63.4\%), which are aligned with our DFAT.

\textbf{Several cues remained salient after five months.}
Five months later, participants still reported most frequently using unnatural facial skin (78.4\%) and overly perfect and idealized style (74.3\%) as cues. Many participants also retained some training memory, with 62.3\% remembering at least some content and 26.9\% remembering most of it or very well. Among those who described what they remembered, 73.9\% mentioned artifact cues, 65.1\% mentioned global or style cues, and 42.6\% mentioned local or skin-related cues. These responses suggest that the effect of the DFAT-guided training is sustainable over a relatively long period of time. 

\textbf{Training was described as clear, useful, and informative.}
After excluding empty entries and replies such as ``N/A'' or ``nothing'', two co-authors conducted keyword-based content analysis on 147 main-study and 97 follow-up responses. In the main study, 59.2\% of the responses were positive or described the study as clear, and 25.9\% described the training as useful or informative. Another 13.6\% pointed to the difficulty of the task, often noting that the face images appeared realistic. A similar pattern was observed in the follow-up study: 45.4\% were positive about the training and its clarity, and 9.3\% mentioned realistic-looking images and difficulty of distinguishing them from real faces. Overall, the feedback proves that our deepfake training materials were well-received and valued by the participants.

\section{Discussion}
\label{sec:further_discussions}

\subsection{Key Findings}

\textbf{Human performance in detecting deepfakes differs across generative models.} Existing literature shows that untrained users often struggle to identify high-quality deepfake faces, with several GAN-based studies reporting near-chance performance~\cite{Frank-DeepfakeHumanDetection-SP2024, Shen:study2021, Cooke:Asgood2025}. These findings suggest that deepfake detection is challenging for humans. Our study revealed two key insights: this difficulty varies across generative models and deepfakes generated by more recent and advanced models may not necessarily be harder to detect. Such results can be well explained based on our DFAT, which identifies distinct visible artifacts produced by the generative models that show varying levels of human-recognizability. Local and subtle artifacts are especially difficult for untrained users to notice, resulting in low detection performance for relevant models as reported in prior works. This finding contributes to the existing literature and highlights the need for cross-model evaluation of human performance in detecting deepfakes.

\textbf{Model-specific deepfake training produces better outcomes than general-purpose training.} It is known that feedback, tutorials, AI support, and artifact-based guidance can improve detection, although their effects vary across settings~\cite{Boyd:value2023, Hulzebosch:DetectingCNN-generated2020, Diel:Inabilitytodetect2024, Nightingale-Farid:PNAS2022, Bray-testinghuman-JC2023}. Recent studies have also shown that human-recognizable artifacts in AI-generated images can be identified and categorized~\cite{Negar:AI-Generated:2024, Kamali:Photorealism:CHI2025}. Our results reveal that matching training content with the target generative model produces better training outcomes. This is likely due to the fact that the general-purpose training included mixed artifacts from different models, potentially increasing participants' cognitive load and undermining the training effectiveness. This finding provides practical guidance for future development of model-specific deepfake training interventions.

\textbf{Artifact-based training has lasting value.} Prior studies involving deepfake training mainly evaluated immediate post-training effects~\cite{Alexander-human-CHBR2024}. It remains unclear whether users retain learned artifact cues and maintain improved detection performance over a longer period of time. Our follow-up study addresses this gap: after five months, detection accuracy remained stable, while confidence level showed only limited decline. Participant feedback also shows that many users could still recall and use the artifact cues learned during the training five months ago. These results suggest that artifact-based training can yield sustainable improvement in deepfake detection.

\subsection{Practical Implications}

\textbf{User training remains an important layer of defense.} Platforms can use automated deepfake detectors to identify likely AI-generated face images and display labels or warnings to users. Prior work shows that labels are not always effective; their effect depends on design, user trust, and user interpretation~\cite{Gamage:LabelingSyntheticContent:CHI2025, Holtervennhoff:AILabels:CHI2026}. In practice, images are often screenshotted, cropped, compressed, or reposted to other platforms. During this process, labels may be lost, and image sources may become hard to verify. Users still need to judge whether online images are trustworthy on social networks, often with limited support to verify the source or context~\cite{Geeng:HowPeople2020}. Our work shows that user training can complement platform defenses by helping users make more evidence-based judgments when external cues are missing.

\textbf{DFAT provides concrete inspection guidance.} Many users know that AI images may contain errors, yet they lack clear guidance on where to look and how to interpret visible anomalies. DFAT gives concrete inspection dimensions for training: error-prone regions, local texture, overall style, and contextual consistency. These dimensions allow training materials to move beyond general warnings and point users to specific visual cues. This structure makes deepfake training easier to teach and easier for non-expert users to remember. It also aligns with recent work showing that targeted deepfake literacy interventions can improve users' ability to reason about manipulated media~\cite{Geissler:DeepfakeLiteracy:CHI2026}. DFAT therefore offers a practical way to organize platform prompts, educational materials, and content moderation guidelines.

\textbf{Model-specific training can reduce misjudgment.} Different generative models produce different types of visual artifacts. Overly broad training may lead users to treat a small set of common artifacts as general signs of AI generation, which can increase mistakes on real images. Model-specific training helps users focus on more relevant visual cues and allows training materials to be updated as generative models change. This matters for deployment because user education should raise awareness of fake images while limiting unnecessary suspicion of real ones. Prior work on uncertainty and confidence in decision support also shows that user-facing information can shape trust and confidence under uncertain judgments~\cite{Zhou:effectsUncertainty2017, Arshad:ConfidenceUncertainty2015}. DFAT supports this goal by grounding user judgment in concrete model-specific visual evidence.

\subsection{Extending DFAT}

\textbf{DFAT can help estimate detection difficulty across models.} Different generative models leave different visual traces, which can affect how easily users can detect them. Prior work showed that artifact types and photorealism shape human detection of AI-generated images~\cite{Kamali:Photorealism:CHI2025}. Our results are consistent with this pattern: models with clear global style problems were easier to detect without training, while models whose errors mainly appeared in local texture required training to educate users about them. This suggests that DFAT can help researchers use artifact patterns to form early expectations about which models may be easier or harder for users to detect and what artifacts may be easier or harder to learn. These expectations still need validation through user studies, and actual human performance should be confirmed in real detection tasks.

\textbf{DFAT can serve as an updatable analysis framework.} Generative models are constantly evolving. Recent work has shown new methods for removing visual imperfections and improving scene consistency in AI-generated images~\cite{Yu:visualGenerated2025, Yuan:GenerativePhotography2025}. A fixed list of artifacts will become outdated quickly. DFAT is designed to be an easily extensible taxonomy. As new models appear, researchers can add new artifact types, refine existing categories, and update user training accordingly. In this way, DFAT can grow from a one-time tool into a dynamic framework for future research and education activities. Future work can also build DFAT-guided training into interactive modules or assistive interfaces that guide users during image inspection~\cite{Poretski:GameLearnability2022, Schemmer:ExplainableAIDecision2022}.

\subsection{Limitations}

While our work studied five generative models, AI-based image generation techniques continue to change quickly, requiring constant updates to the DFAT and training materials. Future work should examine more models and track how artifacts change as generative methods evolve.

Emphasizing controlled comparisons, our online experiment did not cover real-world use cases. To ensure generalizability, larger and more diverse samples are warranted. Although the single-image and side-by-side tasks helped isolate user detection behaviors, face images often appear in social feeds, messaging threads, or news stories. In these settings, users may be affected by context, purpose, and time pressure, which call for more complicated experiment designs in future research. Participant dropout may also introduce selection bias into the five-month retention findings.

DFAT currently focuses on fully generated, photorealistic face images, because this type of deepfakes is among the most abused ones in real-world environment affecting identity trust, social interactions, and online safety. Future research is encouraged to investigate other types of deepfake images and media types as the artifacts involved will differ and the effects of training may change. To facilitate future research, DFAT can be easily extended to support other generative models and incorporate corresponding artifacts across diverse deepfake detection tasks~\cite{Alexander-human-CHBR2024, Wang:DeepfakeDetectionSurvey2025}.

\section{Conclusion}

In this paper, we conducted a comprehensive evaluation of human performance in detecting deepfake face images across five representative generative models following three different architectures. To systematically characterize visual anomalies, we introduced the Deepfake Face Artifact Taxonomy (DFAT), based on which we designed two types of training interventions (general-purpose and model-specific). Our online experiment produced new insights on human performance being strongly model-dependent and images generated by some more recent and advanced models being easier to detect by humans than their older less advanced versions. We further demonstrated that DFAT-guided training materials significantly improved detection accuracy, especially when they were model-specific. The improved accuracy did not come with decreased user confidence or increased stress. We also observed that such training effects persisted after five months, indicating strong sustainability of the training interventions.

Based on the empirical evidence reported, we argue that efforts to raise public awareness should shift from generic alerts to concrete, model-specific and artifact-focused interventions. As deepfake models continue to correct visual imperfections, visual inspection alone may not suffice. Instead of relying solely on human perception, future defenses should integrate technical scaffolding in the human detection process. We advocate implementing a ``Reference Mandate'' in user interfaces that provides authentic baselines for side-by-side comparison. We also encourage the development of active web browser and app plugins that can guide user attention to harder-to-recognize (easier-to-miss) anomalies. As a promising approach to complementing technical solutions, taxonomy-guided public education offers a path to securing the information ecosystem against increasingly hyper-realistic deepfakes.


\section*{Acknowledgments}

This work was partly supported by the National Natural Science Foundation of China (grant numbers 62302427 and 62462060), China's Central Government Guides Local Science and Technology Development Fund (grant number ZYYD2026ZY21), and the Tianshan Talent Training Program (grant number 2022TSYCLJ0036). We also thank the five annotators for contributing to DFAT annotations.

\bibliographystyle{IEEEtran}
\bibliography{main_camera_ready}

@misc{WhichFaceIsReal,
  author = {West, Jevin and Bergstrom, Carl},
  title = {Which Face Is Real?},
  year = {2019},
  url = {https://www.whichfaceisreal.com/},
  note = {University of Washington, Calling Bullshit project. Accessed: 2025-12-18}
}

@inproceedings{Geissler:DeepfakeLiteracy:CHI2026,
    author = {Geissler, Dominique and Robertson, Claire and Feuerriegel, Stefan},
    title = {Designing Effective Digital Literacy Interventions for Boosting Deepfake Discernment},
    year = {2026},
    publisher = {Association for Computing Machinery},
    booktitle = {Proceedings of the 2026 CHI Conference on Human Factors in Computing Systems},
    articleno = {959},
    numpages = {45},
    url = {https://doi.org/10.1145/3772318.3790428},
    doi = {10.1145/3772318.3790428},
}

@inproceedings{Li:AIGNCII:CHI2026,
    author = {Li, Tongxin and Reyes, Katelyn M and Jimenez, Liezeil and Nam, Katie S and Wohn, Donghee Yvette},
    title = {Understanding Educators’ Perceptions of AI-generated Non-consensual Intimate Imagery},
    year = {2026},
    publisher = {Association for Computing Machinery},
    booktitle = {Proceedings of the 2026 CHI Conference on Human Factors in Computing Systems},
    articleno = {536},
    numpages = {17},
    url = {https://doi.org/10.1145/3772318.3790584},
    doi = {10.1145/3772318.3790584},
}

@techreport{Negar:AI-Generated:2024,
    title = "How to Distinguish AI-Generated Images from Authentic Photographs",
    author = "Negar Kamali and Karyn Nakamura and Angelos Chatzimparmpas and Jessica Hullman and Matthew Groh",
    year = "2024",
    month = jun,
    day = "12",
    type = "WorkingPaper",
    institution={Northwestern University},
    doi = {10.48550/arXiv.2406.08651},
    url = {https://doi.org/10.48550/arXiv.2406.08651},
}

@inproceedings{kaushik2021know,
  title = {``{H}ow {I} Know For Sure'': People's Perspectives on Solely Automated Decision-Making ({SADM})},
  author = {Kaushik, Smirity and Yao, Yaxing and Dewitte, Pierre and Wang, Yang},
  booktitle = {Proceedings of the 17th Symposium on Usable Privacy and Security},
  pages = {159--180},
  year = {2021},
  url={https://www.usenix.org/conference/soups2021/presentation/kaushik},
  publisher={USENIX},
}

@article{berens2024taking,
  title={Taking 5 minutes protects you for 5 months: Evaluating an anti-phishing awareness video},
  author={Berens, Benjamin M. and Mossano, Mattia and Volkamer, Melanie},
  journal={Computers \& Security},
  volume={137},
  articleno={103620},
  numpages={19},
  pages={103620:1--103620:19},
  year={2024},
  publisher={Elsevier},
  url={https://doi.org/10.1016/j.cose.2023.103620},
  doi={10.1016/j.cose.2023.103620},
}

@article{Zhang:Subjective2023,
  title = {Subjective and Objective Quality Assessment for in-the-Wild Computer Graphics Images},
  author={Zhang, Zicheng and Sun, Wei and Zhou, Yingjie and Jia, Jun and Zhang, Zhichao and Liu, Jing and Min, Xiongkuo and Zhai, Guangtao},
  journal={ACM Transactions on Multimedia Computing, Communications and Applications},
  volume={20},
  number={4},
  articleno={96},
  numpages={22},
  pages={96:1--96:22},
  year={2023},
  publisher={ACM},
  url = {https://doi.org/10.1145/3631357},
  doi = {10.1145/3631357},
}

@inproceedings{Li:Quality2021,
    author = {Li, Yang and Wang, Shiqi and Zhang, Xinfeng and Wang, Shanshe and Ma, Siwei and Wang, Yue},
    title = {Quality Assessment of End-to-End Learned Image Compression: The Benchmark and Objective Measure},
    year = {2021},
    booktitle = {Proceedings of the 29th ACM International Conference on Multimedia},
    pages = {4297--4305},
    publisher = {ACM},
    url = {https://doi.org/10.1145/3474085.3475569},
    doi = {10.1145/3474085.3475569},
}

@inproceedings{Mohammadi:Subjective2023,
    author = {Mohammadi, Shima and Ascenso, Jo\~{a}o},
    title = {Predictive Sampling for Efficient Pairwise Subjective Image Quality Assessment},
    year = {2023},
    booktitle = {Proceedings of the 31st ACM International Conference on Multimedia},
    pages = {6676--6684},
    numpages = {9},
    publisher = {ACM},
    url = {https://doi.org/10.1145/3581783.3612443},
    doi = {10.1145/3581783.3612443},
}

@inproceedings{Chen:Empowering:ACMCCS2025,
  title={Empowering Parents to Support Children's Online Security and Privacy: Findings from a Randomized Controlled Trial},
  author={Chen, Xiaowei and Distler, Verena and Gordon, Chloe and Yao, Yaxing and Teuber, Ziwen},
  booktitle={Proceedings of the 2025 ACM SIGSAC Conference on Computer and Communications Security},
  pages={1934--1948},
  year={2025},
  doi = {10.1145/3719027.3765214},
  url = {https://doi.org/10.1145/3719027.3765214},
    publisher = {ACM},
}

@article{Vaccari-Chadwick:DeepfakeDisinfo:SMS2020,
author = {Cristian Vaccari and Andrew Chadwick},
title ={Deepfakes and Disinformation: Exploring the Impact of Synthetic Political Video on Deception, Uncertainty, and Trust in News},
journal = {Social Media + Society},
volume = {6},
number = {1},
articleno = {2056305120903408},
numpages = {13},
year = {2020},
doi = {10.1177/2056305120903408},
url = {https://doi.org/10.1177/2056305120903408},
publisher = {SAGE},
}

@article{Furizal:DeepfakePorn:SSHC2025,
author = {Furizal and Alfian Ma'arif and Hari Maghfiroh and Iswanto Suwarno and Denis Prayogi and  Kariyamin and Syahrani Lonang and Abdel-Nasser Sharkawy},
title = {Social, legal, and ethical implications of AI-Generated deepfake pornography on digital platforms: A systematic literature review},
journal = {Social Sciences \& Humanities Open},
volume = {12},
articleno = {101882},
numpages = {25},
pages = {101882:1--101882:25},
year = {2025},
doi = {10.1016/j.ssaho.2025.101882},
url = {https://doi.org/10.1016/j.ssaho.2025.101882},
publisher = {Elsevier},
}

@article{Alexander:DeepfakeCyberbullying:CH2025,
author = {Sergio Alexander},
title = {Deepfake Cyberbullying: The Psychological Toll on Students and Institutional Challenges of AI-Driven Harassment},
journal = {The Clearing House: A Journal of Educational Strategies, Issues and Ideas},
volume = {98},
number = {2},
pages = {36--50},
year = {2025},
doi = {10.1080/00098655.2025.2488777},
url = {https://doi.org/10.1080/00098655.2025.2488777},
publisher = {Routledge},
}

@article{Lin:DeepfakeFraud:FiL2025,
author={Lin, Leo S. F.},
title={Examining the Role of Deepfake Technology in Organized Fraud: Legal, Security, and Governance Challenges},
journal={Frontiers in Law},
volume={4},
year={2025},
pages={6--17},
doi={10.6000/2817-2302.2025.04.02},
url={https://doi.org/10.6000/2817-2302.2025.04.02},
publisher={Lifescience Global},
}

@article{Mirsky-Lee:DeepfakeSurvey2021,
author = {Mirsky, Yisroel and Lee, Wenke},
title = {The Creation and Detection of Deepfakes: A Survey},
journal = {ACM Computing Surveys},
year = {2021},
volume = {54},
number = {1},
articleno = {7},
numpages = {41},
pages = {7:1--7:41},
doi = {10.1145/3425780},
url = {https://doi.org/10.1145/3425780},
publisher = {ACM},
}

@article{Dixit:DeepfakeAudioSurvey2023,
author = {Dixit, Abhishek and Kaur, Nirmal and Kingra, Staffy},
title = {Review of audio deepfake detection techniques: Issues and prospects},
journal = {Expert Systems},
volume = {40},
number = {8},
articleno = {e13322},
numpages = {19},
pages = {e13322:1--e13322:19},
year = {2023},
doi = {10.1111/exsy.13322},
url = {https://doi.org/10.1111/exsy.13322},
publisher = {John Wiley \& Sons, Inc},
}

@article{Li:DeepfakeSpeechSurvey2025,
author = {Li, Menglu and Ahmadiadli, Yasaman and Zhang, Xiao-Ping},
title = {A Survey on Speech Deepfake Detection},
journal = {ACM Computing Surveys},
year = {2025},
volume = {57},
number = {7},
articleno = {165},
numpages = {38},
pages = {165:1--165:38},
doi = {10.1145/3714458},
url = {https://doi.org/10.1145/3714458},
publisher = {ACM},
}

@INPROCEEDINGS{Pu:DeepfakeTextDetection:SP2023,
  author={Pu, Jiameng and Sarwar, Zain and Abdullah, Sifat Muhammad and Rehman, Abdullah and Kim, Yoonjin and Bhattacharya, Parantapa and Javed, Mobin and Viswanath, Bimal},
  title={Deepfake Text Detection: Limitations and Opportunities}, 
  booktitle={Proceedings of the 2023 IEEE Symposium on Security and Privacy}, 
  year={2023},
  pages={1613--1630},
  doi={10.1109/SP46215.2023.10179387},
  url={https://doi.org/10.1109/SP46215.2023.10179387},
  publisher={IEEE},
}

@inproceedings{Uchendu:DeepfakeTextsTutorial:NAACL2024,
    title = {Catch Me If You {GPT}: Tutorial on Deepfake Texts},
    author = {Uchendu, Adaku  and Venkatraman, Saranya  and Le, Thai  and Lee, Dongwon},
    booktitle = {Proceedings of the 2024 Conference of the North American Chapter of the Association for Computational Linguistics: Human Language Technologies (Volume 5: Tutorial Abstracts)},
    year = {2024},
    publisher = {ACL},
    doi = {10.18653/v1/2024.naacl-tutorials.1},
    url = {https://doi.org/10.18653/v1/2024.naacl-tutorials.1},
    pages = {1--7},
}

@inproceedings{Caffagni:MLLMsurvey2024,
  title={The Revolution of Multimodal Large Language Models: A Survey},
  author={Caffagni, Davide and Cocchi, Federico and Barsellotti, Luca and Moratelli, Nicholas and Sarto, Sara and Baraldi, Lorenzo and Cornia, Marcella and Cucchiara, Rita},
  booktitle = {Findings of the Association for Computational Linguistics: ACL 2024},
  year={2024},
  pages = {13590--13618},
  doi = {10.18653/v1/2024.findings-acl.807},
  url = {https://doi.org/10.18653/v1/2024.findings-acl.807},
    publisher = {ACL},
}

@ARTICLE{Zhang:LVMsurvey2024,
  author={Zhang, Jingyi and Huang, Jiaxing and Jin, Sheng and Lu, Shijian},
  title={Vision-Language Models for Vision Tasks: A Survey}, 
  journal={IEEE Transactions on Pattern Analysis and Machine Intelligence}, 
  year={2024},
  volume={46},
  number={8},
  pages={5625--5644},
  doi={10.1109/TPAMI.2024.3369699},
  url={https://doi.org/10.1109/TPAMI.2024.3369699},
  publisher={IEEE},
}

@article{Yin:MLLMsurvey2024,
    author = {Yin, Shukang and Fu, Chaoyou and Zhao, Sirui and Li, Ke and Sun, Xing and Xu, Tong and Chen, Enhong},
    title = {A survey on multimodal large language models},
    journal = {National Science Review},
    volume = {11},
    number = {12},
    articleno = {nwae403},
	numpages = {20},
	pages = {nwae403:1--nwae403:20},
    year = {2024},
    doi = {10.1093/nsr/nwae403},
    url = {https://doi.org/10.1093/nsr/nwae403},
    publisher = {Oxford University Press},
}

@ARTICLE{Wang:DeepfakeDetectionSurvey2025,
  author={Wang, Junke and Li, Zhenxin and Zhang, Chao and Chen, Jingjing and Wu, Zuxuan and Davis, Larry S. and Jiang, Yu-Gang},
  title={Fighting Malicious Media Data: A Survey on Tampering Detection and Deepfake Detection}, 
  journal={Proceedings of the IEEE}, 
  year={2025},
  volume={113},
  number={3},
  pages={287--311},
  doi={10.1109/JPROC.2025.3576367},
  url={https://doi.org/10.1109/JPROC.2025.3576367},
  publisher={IEEE},
}

@misc{Midjourney,
  title        = {{Midjourney}},
  howpublished = {Online model},
  url          = {https://www.midjourney.com/},
  year         = {2025},
  note         = {Accessed: 2025-11-17},
}

@misc{Diel:Inabilitytodetect2024,
  title={Inability to detect deepfakes: Deepfake detection training improves detection accuracy, but increases emotional distress and reduces self-efficacy},
  author={Diel, Alexander and B{\"a}uerle, Alexander and Teufel, Martin},
  howpublished={Preprint},
  year={2024},
  doi={10.31219/osf.io/muwnj},
  url={https://doi.org/10.31219/osf.io/muwnj},
}

@article{Boyd:value2023,
  title={The Value of {AI} Guidance in Human Examination of Synthetically-Generated Faces},
  author={Boyd, Aidan and Tinsley, Patrick and Bowyer, Kevin and Czajka, Adam},
  journal={Proceedings of the AAAI Conference on Artificial Intelligence},
  volume={37},
  number={5},
  pages={5930--5938},
  year={2023},
  doi={10.1609/aaai.v37i5.25734},
  url={https://doi.org/10.1609/aaai.v37i5.25734},
  publisher={AAAI},
}

@article{Kammoun:GANsurvey2022,
title={Generative Adversarial Networks for Face Generation: A Survey},
  author={Kammoun, Amina and Slama, Rim and Tabia, Hedi and Ouni, Tarek and Abid, Mohmed},
  journal={ACM Computing Surveys},
  volume={55},
  number={5},
  articleno={94},
  numpages={37},
  pages={94:1--94:37},
  year={2022},
  publisher={ACM},
  url = {https://doi.org/10.1145/3527850},
  doi = {10.1145/3527850},
}

@article{Yu:visualGenerated2025,
  title={{Yuan}: Yielding Unblemished Aesthetics Through a Unified Network for Visual Imperfections Removal in Generated Images},
  author={Yu, Zhenyu and Chan, Chee Seng},
  journal={Proceedings of the AAAI Conference on Artificial Intelligence},
  volume={39},
  number={9},
  pages={9716--9724},
  year={2025},
  url = {https://doi.org/10.1609/aaai.v39i9.33053},
  doi = {10.1609/aaai.v39i9.33053},
  publisher={AAAI},
}

@inproceedings{Yuan:GenerativePhotography2025,
    author    = {Yuan, Yu and Wang, Xijun and Sheng, Yichen and Chennuri, Prateek and Zhang, Xingguang and Chan, Stanley},
    title     = {Generative Photography: Scene-Consistent Camera Control for Realistic Text-to-Image Synthesis},
    booktitle = {Proceedings of the 2025 IEEE/CVF Conference on Computer Vision and Pattern Recognition},
    year      = {2025},
    pages     = {7920--7930},
    publisher ={IEEE},
}

@article{Guo:ldfnet2023,
  title={{LDFnet}: Lightweight Dynamic Fusion Network for Face Forgery Detection by Integrating Local Artifacts and Global Texture Information},
  author={Guo, Zhiqing and Wang, Liejun and Yang, Wenzhong and Yang, Gaobo and Li, Keqin},
  journal={IEEE Transactions on Circuits and Systems for Video Technology},
  volume={34},
  number={2},
  pages={1255--1265},
  year={2023},
  publisher={IEEE},
  doi = {10.1109/TCSVT.2023.3289147},
  url = {https://doi.org/10.1109/TCSVT.2023.3289147},
}

@inproceedings{Poretski:GameLearnability2022,
    author = {Poretski, Lev and Tang, Anthony},
    title = {Press {A} to Jump: Design Strategies for Video Game Learnability},
    year = {2022},
    publisher = {ACM},
    booktitle = {Proceedings of the 2022 CHI Conference on Human Factors in Computing Systems},
    articleno = {155},
    numpages = {26},
    pages = {155:1--155:26},
    url = {https://doi.org/10.1145/3491102.3517685},
    doi = {10.1145/3491102.3517685},
}

@inproceedings{Zhou:effectsUncertainty2017,
  title={Effects of Uncertainty and Cognitive Load on User Trust in Predictive Decision Making},
  author={Zhou, Jianlong and Arshad, Syed Z and Luo, Simon and Chen, Fang},
  booktitle={Human-Computer Interaction -- INTERACT 2017: 16th IFIP TC 13 International Conference, Mumbai, India, September 25-29, 2017, Proceedings, Part IV},
  pages={23--39},
  year={2017},
  organization={Springer},
  url = {https://doi.org/10.1007/978-3-319-68059-0_2},
  doi = {10.1007/978-3-319-68059-0_2},
}

@inproceedings{Arshad:ConfidenceUncertainty2015,
    author = {Arshad, Syed Z. and Zhou, Jianlong and Bridon, Constant and Chen, Fang and Wang, Yang},
    title = {Investigating User Confidence for Uncertainty Presentation in Predictive Decision Making},
    year = {2015},
    publisher = {ACM},
    booktitle = {Proceedings of the 2015 Annual Meeting of the Australian Special Interest Group for Computer Human Interaction},
    pages = {352--360},
    numpages = {9},
    url = {https://doi.org/10.1145/2838739.2838753},
    doi = {10.1145/2838739.2838753},
}

@inproceedings{Geeng:HowPeople2020,
    author = {Geeng, Christine and Yee, Savanna and Roesner, Franziska},
    title = {Fake News on {Facebook} and {Twitter}: Investigating How People (Don't) Investigate},
    year = {2020},
    publisher = {ACM},
    booktitle = {Proceedings of the 2020 CHI Conference on Human Factors in Computing Systems},
    articleno = {655},
    pages = {655:1--655:14},
    numpages = {14},
    doi = {10.1145/3313831.3376784},
    url = {https://doi.org/10.1145/3313831.3376784},
}

@inproceedings{Schemmer:ExplainableAIDecision2022,
author = {Schemmer, Max and Hemmer, Patrick and Nitsche, Maximilian and K\"{u}hl, Niklas and V\"{o}ssing, Michael},
title = {A Meta-Analysis of the Utility of Explainable Artificial Intelligence in Human-AI Decision-Making},
year = {2022},
publisher = {ACM},
booktitle = {Proceedings of the 2022 AAAI/ACM Conference on AI, Ethics, and Society},
pages = {617--626},
doi = {10.1145/3514094.3534128},
url = {https://doi.org/10.1145/3514094.3534128},
}

@inproceedings{Choi:stargan2018,
  title={{StarGAN}: Unified Generative Adversarial Networks for Multi-domain Image-to-Image Translation},
  author={Choi, Yunjey and Choi, Minje and Kim, Munyoung and Ha, Jung-Woo and Kim, Sunghun and Choo, Jaegul},
  booktitle={Proceedings of the 2018 IEEE Conference on Computer Vision and Pattern Recognition},
  pages={8789--8797},
  year={2018},
  publisher={IEEE},
  doi={10.1109/CVPR.2018.00916},
  url={https://doi.org/10.1109/CVPR.2018.00916},
}

@inproceedings{Chen:SimSwap2020,
    author = {Chen, Renwang and Chen, Xuanhong and Ni, Bingbing and Ge, Yanhao},
    title = {{SimSwap}: An Efficient Framework For High Fidelity Face Swapping},
    year = {2020},
    publisher = {ACM},
    booktitle = {Proceedings of the 28th ACM International Conference on Multimedia},
    pages = {2003--2011},
    url = {https://doi.org/10.1145/3394171.3413630},
    doi = {10.1145/3394171.3413630},
}

@inproceedings{Zi:WildDeepfake2020,
author = {Zi, Bojia and Chang, Minghao and Chen, Jingjing and Ma, Xingjun and Jiang, Yu-Gang},
title = {{WildDeepfake}: A Challenging Real-World Dataset for Deepfake Detection},
year = {2020},
publisher = {ACM},
url = {https://doi.org/10.1145/3394171.3413769},
doi = {10.1145/3394171.3413769},
booktitle = {Proceedings of the 28th ACM International Conference on Multimedia},
pages = {2382--2390},
}

@inproceedings{Hulzebosch:DetectingCNN-generated2020,
  title={Detecting {CNN}-Generated Facial Images in Real-World Scenarios},
  author={Hulzebosch, Nils and Ibrahimi, Sarah and Worring, Marcel},
  booktitle={Proceedings of the 2020 IEEE/CVF Conference on Computer Vision and Pattern Recognition Workshops},
  pages={642--643},
  year={2020},
  doi={10.1109/CVPRW50498.2020.00329},
  url={https://doi.org/10.1109/CVPRW50498.2020.00329},
  publisher={IEEE},
}

@article{Lu:seeing2023,
  title={Seeing is not always believing: Benchmarking Human and Model Perception of {AI}-Generated Images},
  author={Lu, Zeyu and Huang, Di and Bai, Lei and Qu, Jingjing and Wu, Chengyue and Liu, Xihui and Ouyang, Wanli},
  journal={Advances in Neural Information Processing Systems},
  volume={36},
  pages={25435--25447},
  year={2023},
}

@inproceedings{Shen:study2021,
  title={A Study of the Human Perception of Synthetic Faces},
  author={Shen, Bingyu and RichardWebster, Brandon and O'Toole, Alice and Bowyer, Kevin and Scheirer, Walter J.},
  booktitle={Proceedings of the 2021 16th IEEE International Conference on Automatic Face and Gesture Recognition},
  numpages={8},
  year={2021},
  publisher={IEEE},
  doi={10.1109/FG52635.2021.9667066},
  url={https://doi.org/10.1109/FG52635.2021.9667066},
}

@article{Cooke:Asgood2025,
  title={As Good as a Coin Toss: Human Detection of {AI}-Generated Content},
  author={Cooke, Di and Edwards, Abigail and Barkoff, Sophia and Kelly, Kathryn},
  journal={Communications of the ACM},
  volume={68},
  number={10},
  pages={100--109},
  year={2025},
  publisher={ACM},
  doi={10.1145/3729417},
  url={https://doi.org/10.1145/3729417},
}

@inproceedings{Frank-DeepfakeHumanDetection-SP2024,
  author={Frank, Joel and Herbert, Franziska and Ricker, Jonas and Schönherr, Lea and Eisenhofer, Thorsten and Fischer, Asja and Dürmuth, Markus and Holz, Thorsten},
  title={A Representative Study on Human Detection of Artificially Generated Media Across Countries},
  booktitle={Proceedings of the 2024 IEEE Symposium on Security and Privacy},
  year={2024},
  pages={55--73},
  doi={10.1109/SP54263.2024.00159},
  url={https://doi.org/10.1109/SP54263.2024.00159},
  publisher={IEEE},
}

@article{Nightingale-Farid:PNAS2022,
author = {Sophie J. Nightingale and Hany Farid},
title = {{AI}-synthesized faces are indistinguishable from real faces and more trustworthy},
journal = {Proceedings of the National Academy of Sciences},
volume = {119},
number = {8},
articleno = {e2120481119},
numpages = {3},
pages = {e2120481119:1--e2120481119:3},
year = {2022},
doi = {10.1073/pnas.2120481119},
url = {https://doi.org/10.1073/pnas.2120481119},
publisher = {National Academy of Sciences},
}

@inproceedings{Karras-StyleGAN2-CVPR2020,
  title={Analyzing and Improving the Image Quality of {StyleGAN}},
  author={Karras, Tero and Laine, Samuli and Aittala, Miika and Hellsten, Janne and Lehtinen, Jaakko and Aila, Timo},
  booktitle={Proceedings of the 2020 IEEE/CVF Conference on Computer Vision and Pattern Recognition},
  pages={8110--8119},
  year={2020},
  doi={10.1109/CVPR42600.2020.00813},
  url={https://doi.org/10.1109/CVPR42600.2020.00813},
  publisher={IEEE},
}

@article{Karras-StyleGAN3-NeurIPS2021,
  title={Alias-Free Generative Adversarial Networks},
  author={Karras, Tero and Aittala, Miika and Laine, Samuli and H{\"a}rk{\"o}nen, Erik and Hellsten, Janne and Lehtinen, Jaakko and Aila, Timo},
  journal={Advances in Neural Information Processing Systems},
  volume={34},
  pages={852--863},
  year={2021},
  url={https://proceedings.neurips.cc/paper/2021/hash/076ccd93ad68be51f23707988e934906-Abstract.html},
}

@inproceedings{Chan-3DGAN-CVPR2022,
  title={Efficient Geometry-aware {3D} Generative Adversarial Networks},
  author={Chan, Eric R. and Lin, Connor Z. and Chan, Matthew A. and Nagano, Koki and Pan, Boxiao and de Mello, Shalini and Gallo, Orazio and Guibas, Leonidas and Tremblay, Jonathan and Khamis, Sameh and Karras, Tero and Wetzstein, Gordon},
  booktitle={Proceedings of the 2022 IEEE/CVF Conference on Computer Vision and Pattern Recognition},
  pages={16123--16133},
  year={2022},
  doi={10.1109/CVPR52688.2022.01565},
  url={https://doi.org/10.1109/CVPR52688.2022.01565},
  publisher={IEEE},
}

@inproceedings{Chai-Anyres-GAN-ECCV2022,
  title={Any-Resolution Training for High-Resolution Image Synthesis},
  author={Chai, Lucy and Gharbi, Michael and Shechtman, Eli and Isola, Phillip and Zhang, Richard},
  booktitle={Computer Vision -- ECCV 2022: 17th European Conference, Tel Aviv, Israel, October 23--27, 2022, Proceedings, Part XVI},
  pages={170--188},
  year={2022},
  doi={10.1007/978-3-031-19787-1_10},
  url={https://doi.org/10.1007/978-3-031-19787-1_10},
  publisher={Springer},
}

@inproceedings{Xiang-GRAM-HD-ICCV2023,
  title={{GRAM-HD}: {3D}-Consistent Image Generation at High Resolution with Generative Radiance Manifolds},
  author={Xiang, Jianfeng and Yang, Jiaolong and Deng, Yu and Tong, Xin},
  booktitle={Proceedings of the 2023 IEEE/CVF International Conference on Computer Vision},
  pages={2195--2205},
  year={2023},
  doi={10.1109/ICCV51070.2023.00209},
  url={https://doi.org/10.1109/ICCV51070.2023.00209},
  publisher={IEEE},
}

@inproceedings{ROmbach-LDM-CVPR2022,
  title={High-Resolution Image Synthesis with Latent Diffusion Models},
  author={Rombach, Robin and Blattmann, Andreas and Lorenz, Dominik and Esser, Patrick and Ommer, Bj{\"o}rn},
  booktitle={Proceedings of the 2022 IEEE/CVF conference on computer vision and pattern recognition},
  pages={10684--10695},
  year={2022},
  doi={10.1109/CVPR52688.2022.01042},
  url={https://doi.org/10.1109/CVPR52688.2022.01042},
  publisher={IEEE},
}

@inproceedings{Karras-FFHQ-CVPR2019,
  title={A Style-Based Generator Architecture for Generative Adversarial Networks},
  author={Karras, Tero and Laine, Samuli and Aila, Timo},
  booktitle={Proceedings of the 2019 IEEE/CVF Conference on Computer Vision and Pattern Recognition},
  pages={4401--4410},
  year={2019},
  doi={10.1109/CVPR.2019.00453},
  url={https://doi.org/10.1109/CVPR.2019.00453},
  publisher={IEEE},
}

@inproceedings{Li-exposing-arxiv2018,
  title={Exposing {DeepFake} Videos by Detecting Face Warping Artifacts},
  author={Li, Yuezun and Lyu, Siwei},
  booktitle={Proceedings of the 2019 IEEE/CVF Conference on Computer Vision and Pattern Recognition Workshops},
  pages={46--52},
  year={2018},
  url={https://openaccess.thecvf.com/content_CVPRW_2019/html/Media_Forensics/Li_Exposing_DeepFake_Videos_By_Detecting_Face_Warping_Artifacts_CVPRW_2019_paper.html},
  publisher={IEEE},
}

@inproceedings{Li-FaceXray-CVPR2020,
  title={Face {X}-Ray for More General Face Forgery Detection},
  author={Li, Lingzhi and Bao, Jianmin and Zhang, Ting and Yang, Hao and Chen, Dong and Wen, Fang and Guo, Baining},
  booktitle={Proceedings of the 2020 IEEE/CVF Conference on Computer Vision and Pattern Recognition},
  pages={5001--5010},
  year={2020},
  doi={10.1109/CVPR42600.2020.00505},
  url={https://doi.org/10.1109/CVPR42600.2020.00505},
  publisher={IEEE},
}

@inproceedings{Zhao-Mat-CVPR2021,
  author={Zhao, Hanqing and Wei, Tianyi and Zhou, Wenbo and Zhang, Weiming and Chen, Dongdong and Yu, Nenghai},
  title={Multi-attentional Deepfake Detection}, 
  booktitle={Proceedings of the 2021 IEEE/CVF Conference on Computer Vision and Pattern Recognition}, 
  year={2021},
  pages={2185--2194},
  doi={10.1109/CVPR46437.2021.00222},
  url={https://doi.org/10.1109/CVPR46437.2021.00222},
  publisher={IEEE},
}

@inproceedings{Qian-F3Net-ECCV2020,
  title={Thinking in Frequency: Face Forgery Detection by Mining Frequency-Aware Clues},
  author={Qian, Yuyang and Yin, Guojun and Sheng, Lu and Chen, Zixuan and Shao, Jing},
  booktitle={Computer Vision -- ECCV 2020: 16th European Conference, Glasgow, UK, August 23--28, 2020, Proceedings, Part XII},
  pages={86--103},
  year={2020},
  doi={10.1007/978-3-030-58610-2_6},
  url={https://doi.org/10.1007/978-3-030-58610-2_6},
  publisher={Springer},
}

@article{Ba-UMFC-AAAI2024,
  title={Exposing the Deception: Uncovering More Forgery Clues for Deepfake Detection},
  author={Ba, Zhongjie and Liu, Qingyu and Liu, Zhenguang and Wu, Shuang and Lin, Feng and Lu, Li and Ren, Kui},
  journal={Proceedings of the AAAI Conference on Artificial Intelligence},
  volume={38},
  number={2},
  pages={719--728},
  year={2024},
  doi={10.1609/aaai.v38i2.27829},
  url={https://doi.org/10.1609/aaai.v38i2.27829},
  publisher={AAAI},
}

@article{Alexander-human-CHBR2024,
  title={Human performance in detecting deepfakes: A systematic review and meta-analysis of 56 papers},
  author={Diel, Alexander and Lalgi, Tania and Schr{\"o}ter, Isabel Carolin and MacDorman, Karl F and Teufel, Martin and B{\"a}uerle, Alexander},
  journal={Computers in Human Behavior Reports},
  volume={16},
  articleno={100538},
  numpages={13},
  pages={100538:1--100538:13},
  year={2024},
  doi={10.1016/j.chbr.2024.100538},
  url={https://doi.org/10.1016/j.chbr.2024.100538},
  publisher={Elsevier},
}

@article{Bray-testinghuman-JC2023,
  title={Testing human ability to detect `deepfake' images of human faces},
  author={Bray, Sergi D. and Johnson, Shane D. and Kleinberg, Bennett},
  journal={Journal of Cybersecurity},
  volume={9},
  number={1},
  articleno={tyad011},
  numpages={18},
  pages={tyad011:1--tyad011:18},
  year={2023},
  publisher={Oxford University Press},
  doi={10.1093/cybsec/tyad011},
  url={https://doi.org/10.1093/cybsec/tyad011},
}

@inproceedings{karras-PGGAN-2018,
    title={Progressive Growing of {GANs} for Improved Quality, Stability, and Variation},
    author={Tero Karras and Timo Aila and Samuli Laine and Jaakko Lehtinen},
    booktitle={Proceedings of the 2018 International Conference on Learning Representations},
    year={2018},
    numpages={26},
    url={https://openreview.net/forum?id=Hk99zCeAb},
}

@inproceedings{Dustin:SDXL2024,
    title={{SDXL}: Improving Latent Diffusion Models for High-Resolution Image Synthesis},
    author={Dustin Podell and Zion English and Kyle Lacey and Andreas Blattmann and Tim Dockhorn and Jonas M{\"u}ller and Joe Penna and Robin Rombach},
    booktitle={Proceedings of the 12th International Conference on Learning Representations},
    year={2024},
    url={https://openreview.net/forum?id=di52zR8xgf},
}

@misc{Ramesh:DALL-E2022,
  title={Hierarchical Text-Conditional Image Generation with CLIP Latents},
  author={Ramesh, Aditya and Dhariwal, Prafulla and Nichol, Alex and Chu, Casey and Chen, Mark},
  howpublished={Preprint arXiv:2204.06125},
  year={2022},
  doi={10.48550/arXiv.2204.06125},
  url={https://doi.org/10.48550/arXiv.2204.06125},
}

@misc{SD3.5-large,
  title        = {{stabilityai/stable-diffusion-3.5-large}},
  author       = {{Stability AI}},
  howpublished = {Online model at Hugging Face},
  url          = {https://huggingface.co/stabilityai/stable-diffusion-3.5-large},
  year         = {2024},
  note         = {Accessed: 2025-08-03},
}

@misc{SD3.5-news,
  title        = {Introducing {Stable Diffusion} 3.5},
  author       = {{Stability AI}},
  howpublished = {News release},
  url          = {https://stability.ai/news/introducing-stable-diffusion-3-5},
  year         = {2024},
  note         = {Accessed: 2025-08-03},
}

@misc{GPT-5,
  title        = {{GPT}-5 is here},
  author       = {{OpenAI}},
  howpublished = {Web page},
  year         = {2025},
  url          = {https://openai.com/gpt-5/},
  note         = {Accessed: 2025-09-07}
}

@inproceedings{Wobbrock:ARTANOVA:CHI2011,
    title = {The Aligned Rank Transform for Nonparametric Factorial Analyses Using Only {ANOVA} Procedures},
    author = {Wobbrock, Jacob O. and Findlater, Leah and Gergle, Darren and Higgins, James J.},
    year = {2011},
    publisher = {ACM},
    booktitle = {Proceedings of the 2011 SIGCHI Conference on Human Factors in Computing Systems},
    pages = {143--146},
    doi = {10.1145/1978942.1978963},
    url = {https://doi.org/10.1145/1978942.1978963},
}

@incollection{Woolson:Wilcoxontest:2007Wiley,
  title={{Wilcoxon} Signed-Rank Test},
  author={Woolson, Robert F.},
  booktitle={Wiley Encyclopedia of Clinical Trials},
  numpages={3},
  year={2007},
  publisher={John Wiley \& Sons, Inc},
  doi = {10.1002/9780471462422.eoct979},
  url = {https://doi.org/10.1002/9780471462422.eoct979},
}

@article{Xu:AdvancementsAIGC:ACMCS2025,
    author = {Xu, Qiang and Mu, Wenpeng and Li, Jianing and Sun, Tanfeng and Jiang, Xinghao},
    title = {Advancements in AI-Generated Content Forensics: A Systematic Literature Review},
    year = {2025},
    issue_date = {February 2026},
    publisher = {Association for Computing Machinery},
    volume = {58},
    number = {3},
    journal = {ACM Comput. Surv.},
    url = {https://doi.org/10.1145/3760526},
    doi = {10.1145/3760526},
}

@inproceedings{Kamali:Photorealism:CHI2025,
    author = {Kamali, Negar and Nakamura, Karyn and Kumar, Aakriti and Chatzimparmpas, Angelos and Hullman, Jessica and Groh, Matthew},
    title = {Characterizing Photorealism and Artifacts in Diffusion Model-Generated Images},
    year = {2025},
    publisher = {Association for Computing Machinery},
    booktitle = {Proceedings of the 2025 CHI Conference on Human Factors in Computing Systems},
    articleno = {794},
    numpages = {26},
    url = {https://doi.org/10.1145/3706598.3713962},
    doi = {10.1145/3706598.3713962},
}

@inproceedings{Gamage:LabelingSyntheticContent:CHI2025,
    author = {Gamage, Dilrukshi and Sewwandi, Dilki and Zhang, Min and Bandara, Arosha K},
    title = {Labeling Synthetic Content: User Perceptions of Label Designs for AI-Generated Content on Social Media},
    year = {2025},
    publisher = {Association for Computing Machinery},
    booktitle = {Proceedings of the 2025 CHI Conference on Human Factors in Computing Systems},
    articleno = {814},
    numpages = {29},
    url = {https://doi.org/10.1145/3706598.3713171},
    doi = {10.1145/3706598.3713171},
}

@inproceedings{Holtervennhoff:AILabels:CHI2026,
    author = {H\"{o}ltervennhoff, Sandra and Ricker, Jonas and Raphael, Maike M. and Schwedes, Charlotte and Weil, Rebecca and Fischer, Asja and Holz, Thorsten and Sch\"{o}nherr, Lea and Fahl, Sascha},
    title = {"That's another doom I haven't thought about": A User Study on AI Labels as a Safeguard Against Image-Based Misinformation},
    year = {2026},
    publisher = {Association for Computing Machinery},
    booktitle = {Proceedings of the 2026 CHI Conference on Human Factors in Computing Systems},
    articleno = {38},
    numpages = {32},
    url = {https://doi.org/10.1145/3772318.3791006},
    doi = {10.1145/3772318.3791006},
}

\appendices

\section{Detailed Review of Prior Related Studies}
\label{app:related_work}

\begin{table*}[!thb]
\centering
\footnotesize
\renewcommand{\arraystretch}{1.1}
\caption{Overview of prior work on human performance in detecting deepfake images, categorized by training strategy.}
\label{tab:related_works}
\begin{tabularx}{\linewidth}{p{2.0cm} l p{2.4cm} p{3.8cm} X}
\toprule
\textbf{Training strategy} & \textbf{Study} & \textbf{Stimulus type(s)} & \textbf{Human performance} & \textbf{Deepfake model(s) (year)}\\
\midrule
\multirow{2}{=}{\textbf{Deepfake knowledge}} & Bray et al.~\cite{Bray-testinghuman-JC2023} & Image (face) & Not significant (avg.~59.7\% to 64.3\%) & StyleGAN2 (2020)\\
\cmidrule(lr){2-5}
& Nightingale et al.~\cite{Nightingale-Farid:PNAS2022} & Image (face) & Improved (avg. 48.2\% to 59.0\%) & StyleGAN2 (2020)\\
\midrule
\multirow{6}{=}{\textbf{Feedback training \& AI support}} & Boyd et al.~\cite{Boyd:value2023} & Image (face) & Improved fake-image accuracy (avg.~51.9\% to 65.2\%); no model-specific accuracy reported & ProGAN (2019), StyleGAN (2019), StyleGAN2 (2020), StyleGAN2-ADA (2020), StyleGAN3 (2021), StarGANv2 (2020)\\
\cmidrule(lr){2-5}
& Hulzebosch et al.~\cite{Hulzebosch:DetectingCNN-generated2020} & Image (face) & Improved (avg. 67.6\% to 72.9\%) & StarGAN (2018), GLOW (2018), ProGAN (2019), StyleGAN (2019)\\
\cmidrule(lr){2-5}
& Diel et al.~\cite{Diel:Inabilitytodetect2024} & Image (face) & Improved (avg. 44.2\% to 64.7\%) & StyleGAN2 (2020)\\
\midrule
\multirow{10}{=}{\textbf{No training}} & Shen et al.~\cite{Shen:study2021} & Image (face) & Avg.~49.1\% & StyleGAN2 (2020)\\
\cmidrule(lr){2-5}
& Frank et al.~\cite{Frank-DeepfakeHumanDetection-SP2024} & Image (face) and other modalities & Avg.~45.9\%--49.9\% & StyleGAN2 (2020)\\
\cmidrule(lr){2-5}
& Lu et al.~\cite{Lu:seeing2023} & Image (face, landscape, plant, animal, object etc.) & Avg.~64.3\%--66.4\% (faces) & Midjourney V5 (2023)\\
\cmidrule(lr){2-5}
& Cooke et al.~\cite{Cooke:Asgood2025} & Image (face, animal, food, landscape) and other modalities & Avg.~46.6\% (faces) & StyleGAN (2019), StyleGAN2 (2020), StyleGAN3 (2021), Stable Diffusion v2.1 (2022), DALL-E 2 (2022)\\
\bottomrule
\end{tabularx}
\end{table*}

Table~\ref{tab:related_works} compares our work with nine closely related empirical studies, which include a photorealistic face-image condition, report quantitative human detection accuracy, and use controlled experimental settings. The table shows that prior work mainly evaluated older GAN-based models or did not report model-specific human performance, and that existing training interventions provide limited evidence on model-specific training or longer-term retention.

\section{Data Quality and Incentives}
\label{app:data_control}

\noindent\textbf{Data quality control.}
To ensure data quality, we applied the following screening criteria:
\begingroup
\setlength{\IEEEilabelindent}{0pt}
\begin{itemize}
\item \textbf{Attention checks}: The main study included four attention-check trials, one in each pre-/post-training $\times$ single-image/side-by-side testing session. The five-month follow-up study included three attention-check trials across its testing sessions. These trials used easy-to-recognize images, and we excluded participants who failed at least two attention checks in either study.

\item \textbf{Time filtering}: We recorded the completion time for the full session. Submissions completed in $<$6 minutes were treated as invalid due to insufficient effort.

\item \textbf{Data integrity}: We removed submissions with missing responses, inconsistent self-reported demographic information, or an unfinished procedure.
\end{itemize}
\endgroup

\begin{table}[!tb]
\centering
\caption{Demographic breakdown of valid main-study participants ($n=524$). Age is also reported as mean (SD).}
\label{tab:demographics}
\begin{tabular}{l l c c}
\toprule
\multicolumn{2}{l}{\textbf{Characteristic}}
& \textbf{Count ($n$)} & \textbf{Percentage (\%)}\\
\midrule
\multirow{2}{*}{\textbf{Gender}}
& \hspace{1em}Female & 267 & 51.0\\
& \hspace{1em}Male & 257 & 49.0\\
\midrule
\multirow{3}{*}{\textbf{Ethnicity}}
& \hspace{1em}White & 198 & 37.8\\
& \hspace{1em}Black & 163 & 31.1\\
& \hspace{1em}Asian & 163 & 31.1\\
\midrule
\multirow{4}{*}{\textbf{Age}}
& \hspace{1em}18--24 years & 197 & 37.6\\
& \hspace{1em}25--54 years & 202 & 38.5\\
& \hspace{1em}55$+$ years & 125 & 23.9\\
\addlinespace[3pt]
& \hspace{1em}\textit{Mean (SD)}
& \multicolumn{2}{c}{\textit{37.41 (16.05)}}\\
\bottomrule
\end{tabular}
\end{table}

After applying these quality-control measures, 524 participants remained for the main-study analysis and 342 for the five-month follow-up analysis.

\noindent\textbf{Participant incentivization.}
Participants received \pounds5.00 for completing the main study and \pounds3.00 for completing the five-month follow-up study, consistent with Prolific payment guidelines. We also awarded performance-based bonuses within each training group and model condition. In the main study, participants in the top 10\% received an additional \pounds10.00, and those ranked between 10\% and 30\% received an additional \pounds5.00. In the follow-up study, participants in the top 10\% received an additional \pounds6.00, and those ranked between 10\% and 30\% received an additional \pounds3.00.

\noindent\textbf{Participant demographics.}
Table~\ref{tab:demographics} summarizes the demographic characteristics of the 524 valid participants.

\begin{figure*}[!tbh]
\centering
\includegraphics[width=0.9\textwidth]{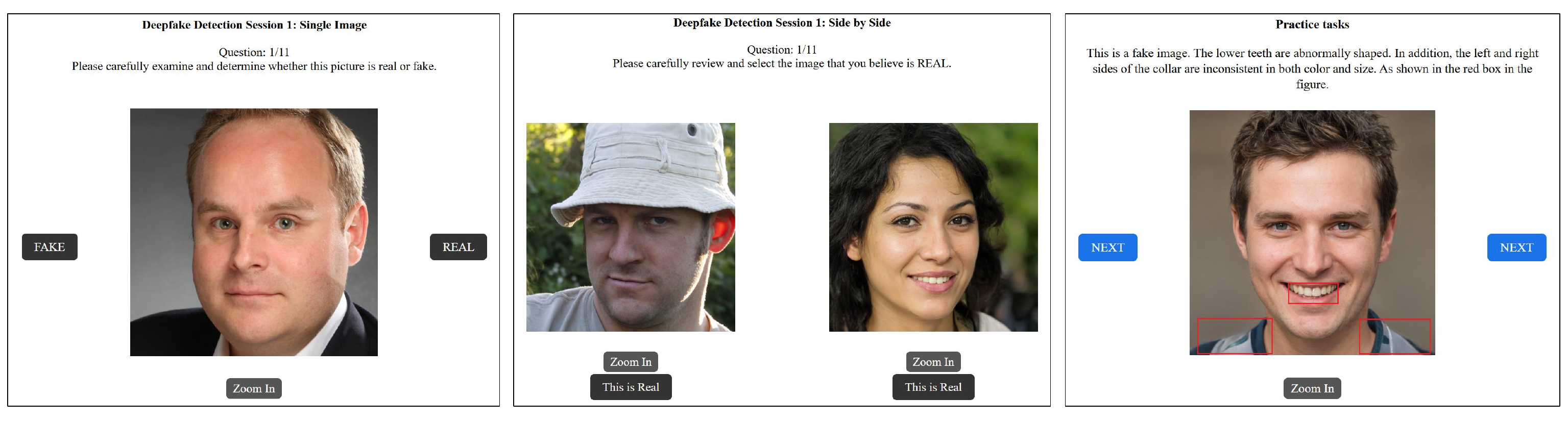}
\caption{Screenshots of the experimental platform testing interfaces.}
\label{fig:platform_screenshots}
\end{figure*}

\section{Platform and Questionnaires}
\label{app:study_materials_questionnaires}

This appendix illustrates the experimental platform and lists the questionnaire items used in the main study and the follow-up study.

\subsection{Platform User Interface}
\label{app:platform_interface}

Figure~\ref{fig:platform_screenshots} shows representative interfaces from the experiment platform, including the single-image test, side-by-side test, and practice-task feedback interfaces. In the practice-task feedback interface, participants received the correct answer and an explanation after submitting their response.

\subsection{Confidence and Perceived-Stress Questions}
\label{app:self_assessment}

After each testing session, participants answered two self-assessment questions about confidence and perceived stress. These questions supported the analyses of subjective states in Section~\ref{subsec:behavior_subjective} and Section~\ref{subsec:retention}. Both used a 5-point Likert scale: \textit{Very low; Low; Moderate; High; Very high}.

\begingroup
\noindent\textbf{Q1.} \textit{Overall, how confident did you feel about your answers in the session you just completed?}

\noindent\textbf{Q2.} \textit{Overall, how stressed did you feel during the session you just completed?}
\endgroup

\subsection{Main-Study Feedback Questionnaire}
\label{app:main_feedback_questions}

The following feedback questions were shown after the post-training tests in the main study.

\begingroup
\noindent\textbf{Q1.} \textit{Do you agree that deepfake technology is a significant societal problem?}

\textit{Options:} Strongly disagree; Disagree; Neutral; Agree; Strongly agree.

\noindent\textbf{Q2.} \textit{Do you agree that the training session was helpful for improving your ability to identify deepfake images?}

\textit{Options:} Strongly disagree; Disagree; Neutral; Agree; Strongly agree.

\noindent\textbf{Q3.} \textit{What aspects of the training do you find particularly useful?}

\textit{Options (select all that apply):} Training video; Practice tasks; Others (open-ended text).

\noindent\textbf{Q4.} \textit{What are the main clues you now look for when identifying deepfake images?}

\textit{Options (select all that apply):} Background problems; Unnatural hair; Unnatural facial skin; Accessory rendering anomalies; Overly perfect and idealized style; Others (open-ended text).

\noindent\textbf{Q5.} \textit{Do you have any other feedback or comments?}

\textit{Response format:} Open-ended text.
\endgroup

\subsection{Follow-Up Feedback Questionnaire}
\label{app:followup_feedback_questions}

The following feedback questions were shown after the five-month follow-up tests.

\begingroup
\noindent\textbf{Q1.} \textit{What are the main clues you look for when identifying deepfake images?}

\textit{Options (select all that apply):} Background problems; Unnatural hair; Unnatural facial skin; Accessory rendering anomalies; Overly perfect and idealized style; Others (open-ended text).

\noindent\textbf{Q2.} \textit{Do you still remember the training content from the previous study?}

\textit{Options:} Not at all; A little; Some of it; Most of it; Very well.

\noindent\textbf{Q3.} \textit{If you remember the training at all, what aspects of it do you still remember?}

\textit{Response format:} Open-ended text.

\noindent\textbf{Q4.} \textit{Do you have any other feedback or comments?}

\textit{Response format:} Open-ended text.
\endgroup

\section{Examples of StyleGAN3 Texture Artifacts}
\label{app:stylegan3_examples}

\begin{figure}[H]
\centering
\includegraphics[width=0.75\linewidth]{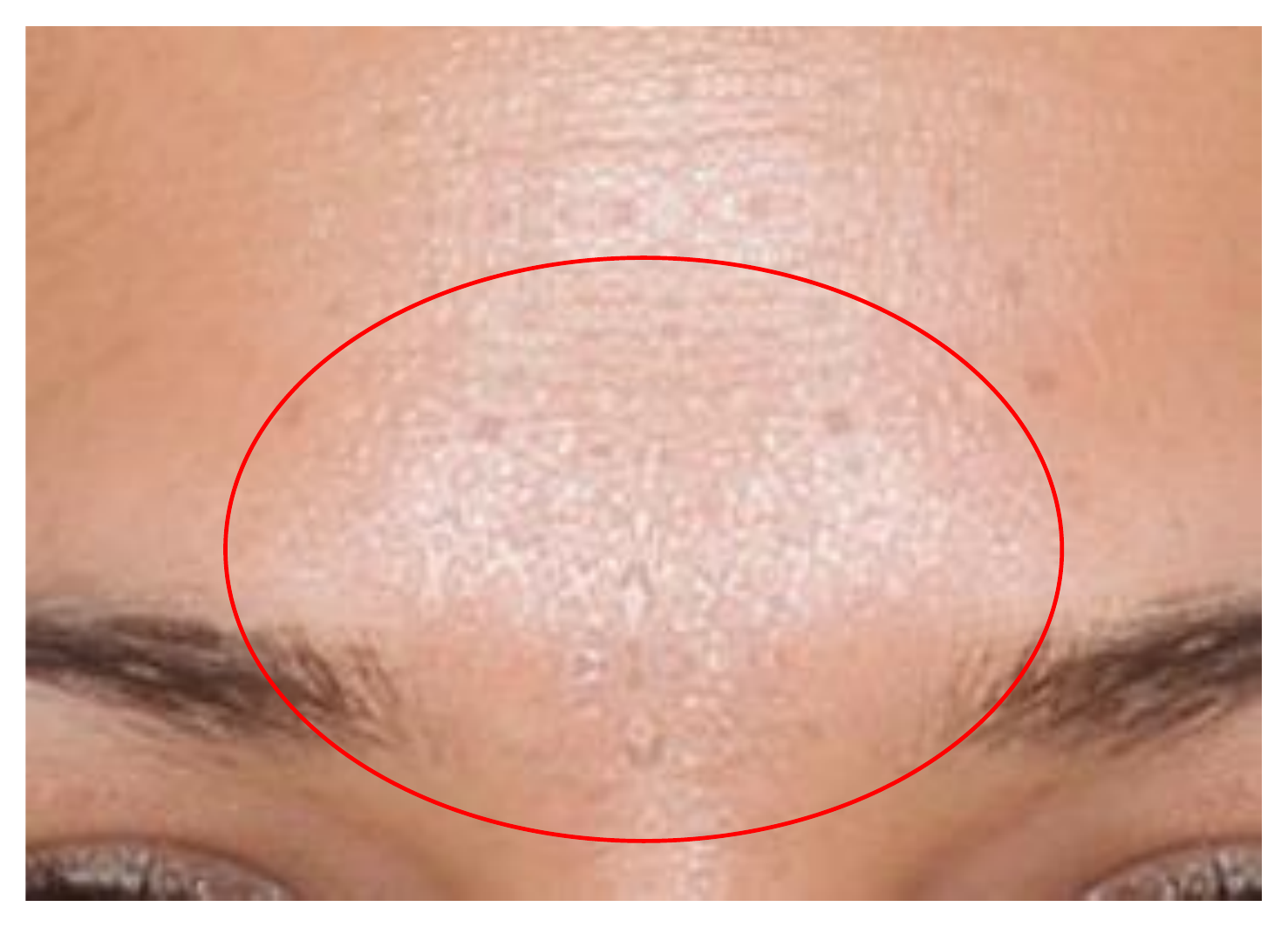}
\caption{Examples of skin texture artifacts in StyleGAN3-generated images.}
\label{fig:stylegan3_artifacts}
\end{figure}

Some model-specific artifacts require close inspection to identify. In StyleGAN3-generated images, we observed recurring abnormal skin-texture patterns. These patterns are subtle at normal viewing size but become visible under magnification. Figure~\ref{fig:stylegan3_artifacts} shows two representative examples. In the zoomed-in crops, the skin surface lacks natural pore structure and instead exhibits structured local patterns with oval or diamond-like shapes. Additional annotated examples are available in the artifact repository.

\section{Artifact Availability}
\label{app:artifact_availability}

The DFAT corpus and annotations, generation prompts, study materials, de-identified data, analysis scripts, and experimental website source code are available at \url{https://doi.org/10.5281/zenodo.22843044}.

\section{Ethics Considerations}

The experiments that recruited human participants from Prolific received ethics approval from the Departmental Ethics Review Group at Cambridge Judge Business School, University of Cambridge, UK (reference number: 25-46). Participants provided informed consent before taking part in the experiments, where they were asked to identify real and deepfake face images. They were compensated for their time at a rate following Prolific recommendations and some of them received additional performance-based bonuses. We screened all face images to exclude inappropriate or disturbing content and minimize potential discomfort during the online experiment. Real images were sourced from the FFHQ dataset. We collected Prolific IDs for payment and follow-up recruitment and removed these identifiers from the publicly released study data. This work did not identify vulnerabilities or involve testing live third-party systems. The DFAT and associated training materials are intended to support education, evaluation, and human-centered security research.

To examine the prevalence of artifact types across deepfake models, five annotators who were not co-authors were recruited to label the deepfake images generated by different models. This task was led by the first two authors, whose institution (Xinjiang University, China) did not require a research ethics review process for such low-risk studies involving human participants. No personal or sensitive data was collected for this task, nor were the participants asked to do heavily repetitive work. All annotators did the work voluntarily and no financial compensation was involved.

\newpage

\section{Meta-Review}

The following meta-review was prepared by the program committee for the 2027 IEEE Symposium on Security and Privacy (S\&P) as part of the review process as detailed in the call for papers.

\subsection{Summary}

This paper presents a multi-month spanning study on human performance in detecting deep-fake images and the efficacy of different types of training in that period. As part of this study, the paper systematically studied the artifacts present in deep-fake images and created a taxonomy of those artifacts. Based on the artifacts, the study then conducted an online user-study ($N=524$) where the study focused on different factors that may influence the training efficacy, such as generalized and specialized training, pre-and-post training tasks, and a follow-up study. Through the study, the paper demonstrated that the taxonomy guided approach for training the users on detecting deep fake images is useful and lasting even after months.

\subsection{Scientific Contributions}

\begin{itemize}
\item Creates a New Tool to Enable Future Science
\item Provides a New Data Set For Public Use
\item Provides a Valuable Step Forward in an Established Field
\end{itemize}

\subsection{Reasons for Acceptance}

\begin{enumerate}
\item The paper creates a taxonomy that reflects the large enumeration of artifacts and offers an effective approach for updating training materials, thus enabling future science.

\item The reviewers found the paper well organized, experimentally detailed, and rigorous with human-subject studies, thus providing a valuable step towards making progress in understanding human performance in detecting deep-fake images.

\item The open-sourced artifact is well organized and documented to make it suitable for public use and to support future research.
\end{enumerate}

\end{document}